\documentclass[times, review, 10pt]{elsarticle}
\usepackage[
  top=4.3cm,
  right=4.8cm,
  bottom=4.3cm,
  left=4.8cm
]{geometry}

\usepackage{etoolbox}

\makeatletter
\patchcmd{\pprintMaketitle}
  {\normalsize\elsauthors}
  {\footnotesize\elsauthors}
  {}{}

\patchcmd{\MaketitleBox}
  {\normalsize\elsauthors}
  {\footnotesize\elsauthors}
  {}{}
\makeatother

\usepackage{amssymb}
\usepackage{amsmath}
\usepackage{booktabs}
\usepackage[table]{xcolor}
\usepackage{pifont}
\newcommand{\cmark}{\ding{51}}
\newcommand{\xmark}{\ding{55}}
\usepackage{graphicx}

\def\tsc#1{\csdef{#1}{\textsc{\lowercase{#1}}\xspace}}
\tsc{WGM}
\tsc{QE}
\tsc{EP}
\tsc{PMS}
\tsc{BEC}
\tsc{DE}
\usepackage{multirow}
\journal{New Trend of Multimodal Intelligence: Perception, Computing, Understanding, and Generation via Large AI Models}
\begin{document}
\let\WriteBookmarks\relax
\def\floatpagepagefraction{1}
\def\textpagefraction{.001}
\begin{frontmatter}
\title{\fontsize{14}{16}\selectfont
Prompt-Based Continual Compositional Zero-Shot Learning
\tnotemark[tnote1]}



\affiliation[1]{organization={Department of Artificial Intelligence, Information Technology University},
                addressline={Barki Road}, 
                city={Lahore},
                postcode={54000}, 
                state={Punjab},
                country={Pakistan}}

\author[1]{Sauda Maryam}
\ead{msds22025@itu.edu.pk}

\author[1]{Sara Nadeem}
\ead{phdcs21001@itu.edu.pk}

\author[2]{Faisal Z. Qureshi\corref{cor1}}
\ead{faisal.qureshi@ontariotechu.ca}

\affiliation[2]{
    organization={Department of Computer Science, Ontario Tech University},
    addressline={2000 Simcoe Street North},
    city={Oshawa},
    postcode={L1G 0C5},
    state={Ontario},
    country={Canada}
}
                
\author[1]{Mohsen Ali}
\ead{mohsen.ali@itu.edu.pk}


\begin{abstract}
Continual Compositional Zero-Shot Learning (CCZSL) aims to incrementally learn evolving attribute–object compositions while preserving previously acquired knowledge. Unlike conventional continual learning with disjoint classes, CCZSL is particularly challenging because attributes and objects may reoccur across sessions while compositions remain session-specific, leading to semantic interference, compositional drift, and catastrophic forgetting. Existing CZSL methods are designed for static settings and struggle to balance knowledge retention and adaptation in continual environments.
To address these challenges, we propose Prompt-based Continual Compositional Zero-Shot Learning (PromptCCZSL), a vision–language framework for continual compositional adaptation. PromptCCZSL leverages prompt learning to disentangle compositional and primitive knowledge through session-aware compositional prompts and session-agnostic primitive prompts, enabling a favorable balance between stability and plasticity. Furthermore, we introduce a Cross-Session Knowledge Distillation strategy and a Primitive-aware Anchor Alignment Loss to preserve semantic consistency of previously learned primitives and mitigate representational drift during incremental updates.
To facilitate adaptation to newly introduced concepts while reducing interference, we further employ Orthogonal Projection and Intra-Session Diversity regularization, which encourage discriminative and separable compositional representations.
In addition, we introduce a comprehensive CCZSL evaluation protocol that assesses both catastrophic forgetting and compositional generalization. Extensive experiments on three benchmark datasets demonstrate that PromptCCZSL consistently outperforms existing VLM and non-VLM baselines.
\end{abstract}




\end{frontmatter}
\section{Introduction}
\begin{figure}
    \centering
    \includegraphics[width=0.9\linewidth]{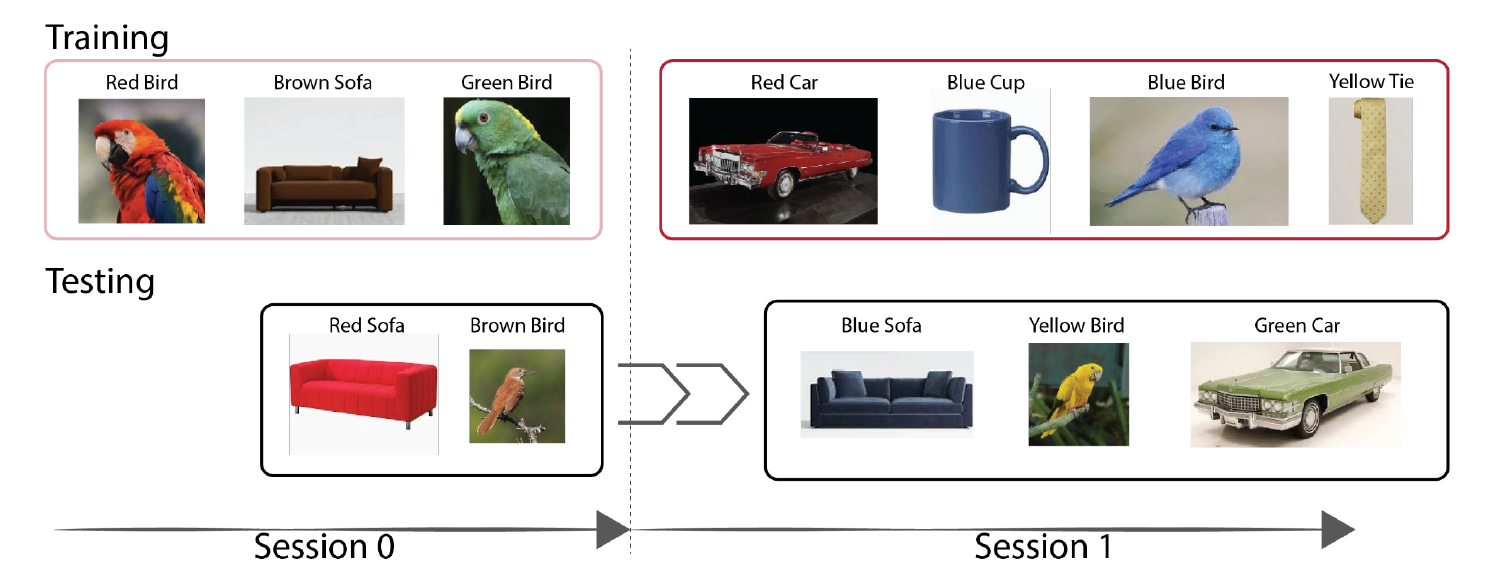}
    \caption{\textbf{Continual Compositional Zero-Shot Learning (CCZSL) setup.} In Session 0, the model learns seen compositions and generalizes to unseen attribute--object pairs. In Session 1, without prior data, it learns new attributes, objects, and compositions while retaining concepts. This enables recognition of new and previously unseen compositions. \textit{Images are from Google Search for educational use.}}
    \label{fig:framework1}
\end{figure}

For reliable scene understanding, models must recognize not only \textit{what} an object is but also \textit{how} it appears (e.g., \textit{wet cat}, \textit{broken glass}), while continuously adapting to newly emerging attribute–object compositions without retraining from scratch.
Conventional Compositional Zero-Shot Learning (CZSL) tackles this problem by decomposing visual concepts into attributes and objects, and recombining seen primitives to recognize unseen zero-shot compositions~\cite{huynh2020compositional, li2022siamese, naeem2021learning, zhang2022learning}.
However, existing CZSL assumes a fixed set of primitives observed during training, limiting its applicability in real-world scenarios where new objects and attributes may appear incrementally. 
In this work, we attempt to address the limitation through Continual Compositional Zero-Shot Learning (CCZSL), which extends CZSL to a continual learning setting in which attribute–object compositions evolve across sequential sessions. An example of CCZSL is shown in Fig.~\ref{fig:framework1}. 
In contrast to standard continual learning with disjoint class sets, CCZSL requires preserving shared primitives while incrementally learning new compositions, making it highly susceptible to catastrophic forgetting and representational drift. 
Specifically, we consider a session-based setting where both compositions and participating primitives may expand over time. For instance, in Session~0 the model is trained on compositions such as red bird, brown sofa, and green bird, and evaluated on unseen compositions such as red sofa and brown bird. In Session~1, new primitives and compositions (e.g., red car, blue cup, blue bird, yellow tie) are introduced without access to previous session data.

This setting introduces two major challenges. First, learning new compositions may degrade performance on previously learned ones due to catastrophic forgetting. Second, shared primitive representations may shift across sessions as the model adapts to new compositions, leading to cross-session representational drift. Together, these effects highlight the stability–plasticity trade-off inherent in CCZSL.
\begin{figure*}
    \centering
    \includegraphics[width=1.0\linewidth]{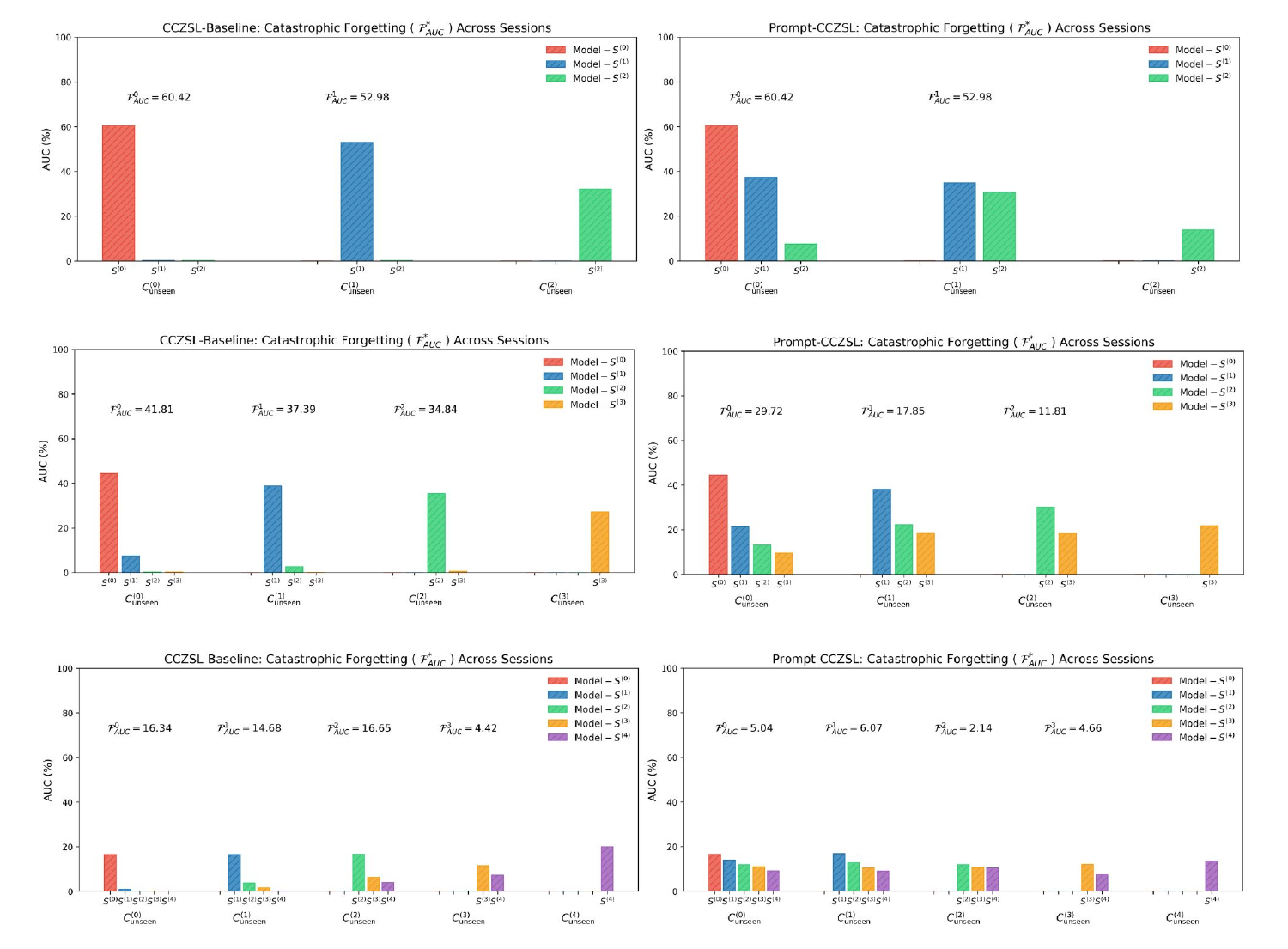}
    \caption{
        \textbf{Catastrophic forgetting in Continual Compositional Zero-Shot Learning (CCZSL) across three benchmarks.}
        Each row corresponds to one dataset (top: UT-Zappos, $T=3$; middle: MIT-States, $T=4$; bottom: CGQA, $T=5$).
        \textbf{Left:} CCZSL-Baseline suffers severe forgetting, where performance on previously unseen compositions deteriorates after learning new sessions.
        \textbf{Right:} PromptCCZSL (ours) consistently preserves higher AUC on previously unseen compositions while adapting to newly introduced compositions, resulting in substantially lower catastrophic forgetting scores, $\mathcal{F}_{\mathrm{AUC}}^{i}$ (Eq.~\ref{eq:avg_fau}), where $i$ indexes the zero-shot compositions evaluated in each incremental session. Models $\mathcal{S}^{(t)}$ denote the model after learning up to session $t$, and are evaluated independently on the zero-shot composition sets from all sessions observed up to $t$. For example, in CGQA, $\mathcal{S}^{(0)},\ldots,\mathcal{S}^{(4)}$ are obtained by sequentially learning $\mathcal{C}_{\mathrm{seen}}^{(0)},\ldots,\mathcal{C}_{\mathrm{seen}}^{(4)}$, and are evaluated independently on the corresponding zero-shot composition sets from sessions $\mathcal{C}_{\mathrm{unseen}}^{(0)},\ldots,\mathcal{C}_{\mathrm{unseen}}^{(4)}$, respectively.
    }
    \label{fig:fauc_all}
\end{figure*}
Recent advances in vision--language models (VLMs), such as CLIP~\cite{radford2021learning}, have demonstrated strong compositional generalization through large-scale contrastive pretraining. Building on this, several works~\cite{lu2023decomposed, nayaklearning, huang2024troika} fine-tune CLIP for compositional understanding using decomposed embeddings and multimodal prompting. On the vision side, the CLIP backbone is typically adapted with parameter-efficient tuning techniques such as LoRA~\cite{hu2022lora} and Adapters~\cite{karimi2021compacter, rebuffi2017learning}, aligning visual and textual representations while preserving pre-trained knowledge. These strategies enable efficient compositional zero-shot transfer and have shown promise in extending the generalization of VLMs to novel compositional scenarios. However, our findings in Fig. \ref{fig:fauc_all} reveal that direct SOTA prompt-based compositional methods~\cite{huang2024troika} in the CCZSL setting (which we define as a baseline) suffer from a fundamental limitation; they are designed under a static compositional assumption, with fixed attribute and object vocabularies. Consequently, they struggle to incorporate new primitives over time, resulting in severe catastrophic forgetting, degraded semantic alignment and representation drift in evolving environments.
Building on these observations, we introduce Prompt-Based Continual Compositional Zero-Shot Learning (PromptCCZSL), a framework that combines vision–language pretraining with continual compositional learning to acquire new attribute–object primitives while preserving prior knowledge.

Unlike prior approaches, PromptCCZSL employs a frozen text backbone with a shared learnable prompt bank, enabling new primitives to be added without updating backbone parameters. Lightweight session-wise visual adapters capture new patterns while reducing catastrophic forgetting, balancing plasticity and stability~\cite{jung2023new}.

Due to limited prior work on adapting CZSL to continual settings, we introduce a student--teacher inspired continual adaptation strategy~\cite{li2017learning}, termed Cross-Session Knowledge Distillation (CSKD), where each session distills knowledge from the preceding model via soft logits. While CSKD preserves historical decision boundaries, our ablations in Tab.~\ref{tab:Table_9} show that distillation alone is insufficient for stable continual learning. We therefore introduce three complementary regularizers: \textit{Primitive-aware Anchor Alignment Loss} (PAAL) for cross-session primitive alignment, \textit{Orthogonal Projection Loss} (OPL) for reducing inter-session interference, and \textit{Intra-session Diversification Loss} (IDL) for improving within-session discrimination.

Finally, we identify representational drift caused by the attention-based composition module of~\cite{huang2024troika}. To mitigate this, we propose \textit{Session-Aware Multimodal Fusion} (SAwM2F), which models session-wise contextual shifts for adaptive composition learning while maintaining semantic consistency across old and new primitives.

The contribution of our work are summarized as follows.
\begin{itemize}
    \item 
    

    We propose \textbf{PromptCCZSL}, a prompt-based continual learning framework that integrates vision-language models for incremental compositional zero-shot learning over attribute--object primitives. We also establish strong prompt-based continual CZSL baselines under a unified training and inference protocol.

    We introduce \textit{Session-Aware Multimodal Fusion} (SAwM2F), which dynamically integrates visual and textual representations to learn session-specific compositions while preserving previously acquired compositional knowledge.
    
    \item We introduce Cross-Session Knowledge Distillation (CSKD) along with primitive-aware Primitive-aware Anchor Alignment (PAAL) for CCZSL that aligns shared attribute–object primitives across sessions, preserving prior compositional knowledge and reducing semantic drift.

     \item 

     We further incorporate \textit{Orthogonal Projection Loss} (OPL) and \textit{Intra-Session Diversification Loss} (IDL) to stabilize compositional embeddings. OPL reduces inter-session interference by separating new primitives from previously learned ones, while IDL promotes intra-session diversity, yielding stable and well-distributed representations.

     \item We establish a Catastrophic Forgetting evaluation protocol (CFZSEval), enabling systematic assessment of compositional generalization and forgetting.

    \item 
    In addition to the widely used UT-Zappos and C-GQA, we extend the CZSL benchmark to the MIT-States dataset under the CCZSL setting. Experiments across all three datasets show that our method consistently outperforms state-of-the-art VLM- and non-VLM-based CZSL approaches, demonstrating strong compositional generalization and robustness in continual learning.
\end{itemize}

\section{Related Work}

\subsection{Compositional Learning}
Compositional Zero-Shot Learning (CZSL) aims to recognize unseen attribute--object pairs using
knowledge from seen compositions. Early work learned joint embeddings for images and compositions~\cite{misra2017red}, modeled attributes as transformations of object representations~\cite{nagarajan2018attributes}, or imposed algebraic constraints to stabilize composition mappings~\cite{li2020symmetry}. ~\cite{jiang2024mutual} improves compositional recognition by modeling the imbalance between attribute and object components and introducing a balancing inductive bias during training. Later methods improved disentanglement via contrastive objectives~\cite{li2022siamese}, and recent visual encoders enhanced object-conditioned attribute reasoning through attention-based disentanglers and conditional attribute modeling~\cite{wang2023learning}. With CLIP-style vision–language models (VLMs), compositional zero-shot learning (CZSL) methods typically freeze the backbone and optimize lightweight adapters or prompt-based modules.  CSP~\cite{nayaklearning} learns attribute and object prompt tokens and composes them within a prompt template for compositional recognition, DFSP~\cite{lu2023decomposed} uses decomposed cross-modal fusion to better align image–text features, Troika~\cite{huang2024troika} models state, object, and composition in parallel using multi-path prompts with cross-modal traction, VisPrompt \cite{shuang2025visual} introduces fine-grained visual-text alignment by treating visual primitives as semantic words and interacting visual attribute and object representations with prompt sub-concepts,, while hierarchical prompt learning [21] aims to reduce semantic gaps between modalities ~\cite{wang2023hierarchical}.However, existing VLM-based CZSL methods remain static and cannot support continual updates of attribute–object compositions.

\subsection{Continual Learning}
Continual learning (CL) learns tasks sequentially while mitigating catastrophic forgetting. Classical CL includes regularization that constrains parameters important to past tasks ~\cite{kirkpatrick2017overcoming}, parameter isolation that allocates task-specific modules or masks~\cite{mallya2018packnet} and replay that rehearses stored exemplars~\cite{isele2018selective}, with recent work improving replay through multi-criteria rehearsal sample selection to enhance the representativeness and discriminability of memory exemplars ~\cite{zhuang2022multi}. Knowledge distillation (KD) plays a central role in CL by transferring knowledge from a frozen teacher model to mitigate forgetting. Learning without Forgetting (LwF) aligns the logits of the current model with those of the teacher to preserve prior knowledge~\cite{li2017learning}. Subsequent methods extend this idea by incorporating feature-level and intermediate representation alignment to further improve stability in class-incremental settings~\cite{jung2023new}. 
Relational KD~\cite{park2019relational} preserves pairwise geometric structure, such as distances and angles to maintain representational consistency. Other approaches directly minimize L2 or MSE losses between feature embeddings or reuse teacher classifiers for stable supervisions~\cite{chen2022knowledge}. Recent methods improve transfer by aligning feature directions while preserving feature magnitudes~\cite{wang2024improving}.  
In Vision Transformers (ViTs), feature distillation depends on where knowledge is transferred.
DeiT~\cite{touvron2021training} introduces a distillation token for teacher–student learning, while ViTKD~\cite{yang2024vitkd} shows that shallow layers are more transferable, while deeper layers are more task-specific.

For large VLMs such as CLIP, parameter-efficient tuning (PEFT) methods freeze the pretrained backbone and learn lightweight modules, such as adapters, LoRA layers, or soft prompts, to preserve zero-shot generalization. LADA~\cite{luo2025lada} uses label-specific adapters and feature distillation with a frozen image encoder to mitigate forgetting. Prompt-based continual learning extends PEFT by learning task-adaptive prompts while keeping the backbone frozen. L2P~\cite{wang2022learning} retrieves task-relevant prompts; DualPrompt~\cite{wang2022dualprompt} separates shared and task-specific prompts; CODA-Prompt~\cite{smith2023coda} dynamically composes decomposed prompt components.

Continual distillation in Vision Transformers improves knowledge retention with efficient inference. KDP~\cite{zhang2024continualprompt} uses learnable distillation prompts with a frozen backbone, while PromptKD~\cite{li2024promptkd} aligns student image logits with teacher text features. Dual-teacher methods~\cite{zheng2024adapt} further combine a frozen foundation model with a task-specific teacher.

\subsection{Continual Compositional Zero-Shot Learning.}
Zero-shot learning (ZSL)~\cite{radford2021learning} transfers knowledge from seen to unseen classes using semantic representations. Compositional zero-shot learning (CZSL) instead targets \textit{combinatorial generalization}, recognizing unseen attribute--object compositions from shared primitives.
Zhang \textit{et al.}~\cite{zhang2024continual} introduced a non-VLM CCZSL framework that incrementally adds attributes and objects using dual KD to preserve prior knowledge. However, it reports only closed-world cumulative AUC and lacks standard CZSL and session-wise forgetting metrics, limiting evaluation of stability--plasticity~\cite{jung2023new} and generalization to new primitives. ~\cite{li2026composition} proposed CompIL, which learns new compositions under a fixed primitive vocabulary using pseudo replay and primitive distillation. In contrast, CCZSL introduces new attributes and objects across sessions, requiring adaptation to both new primitives and unseen compositions.

\begin{figure}
    \centering
    \includegraphics[width=0.9\linewidth]{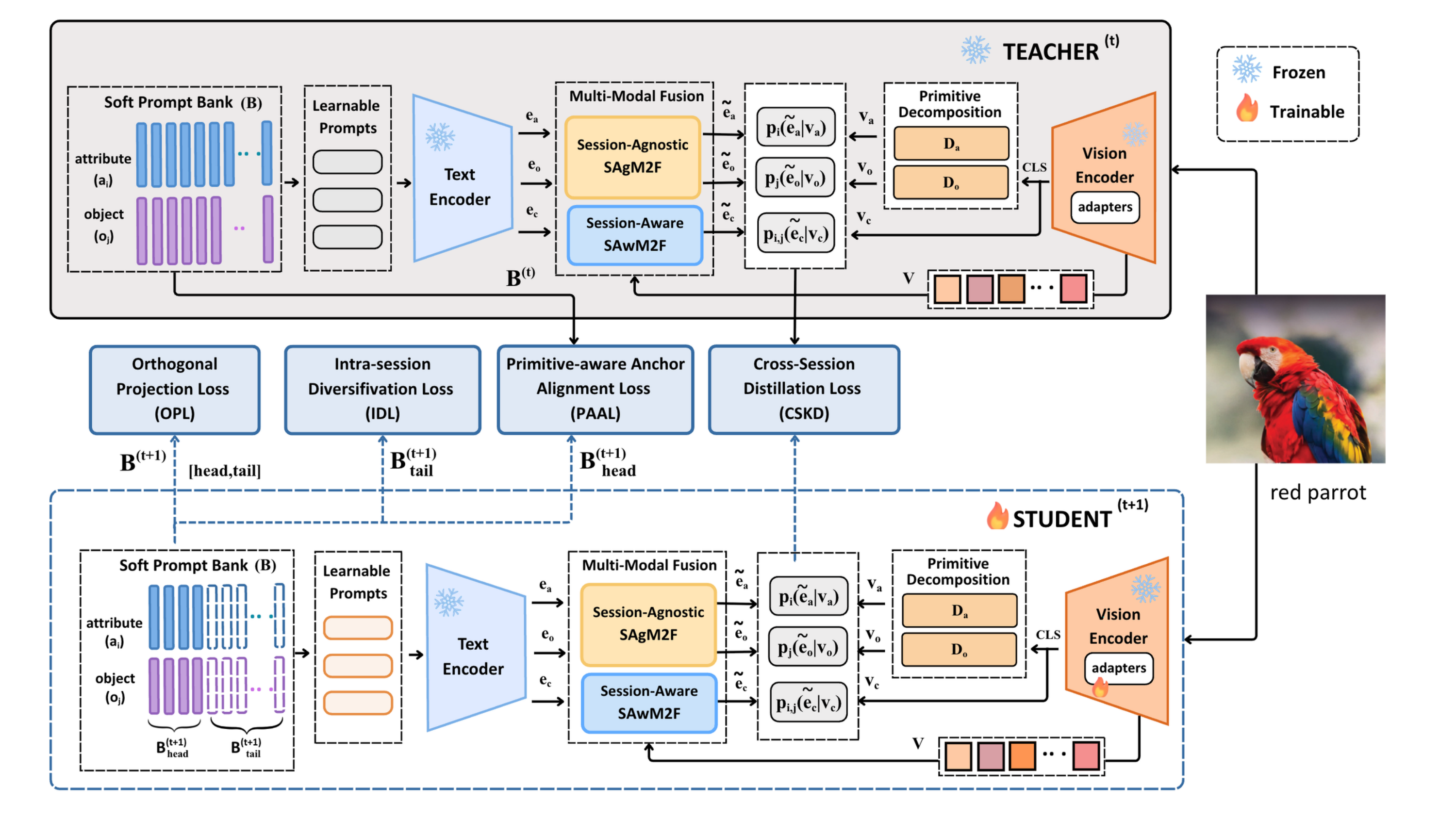}
    \caption{\textbf{Overview of PromptCCZSL.} At session $S^{(t)}$, images are processed using session-specific visual prompts, while the frozen teacher $S^{(t-1)}$ transfers prior knowledge through Cross-Session Knowledge Distillation. Primitive-aware Anchor Loss stabilizes shared primitives, Session-Aware Multi-Modal Fusion refines prompt representations, Orthogonal Projection Loss separates old and new primitives, and Intra-Session Diversification Loss promotes diversity among prompts. Together, these components enable continual adaptation while preserving prior knowledge. (Sec.~\ref{sec:proposed_framework}).
    }
    \label{fig:framework}
\end{figure} 
\section{Priliminaries}
\newcommand{\comp}{\mathcal{C}}
\newcommand{\compSeen}{\comp_\text{seen}}
\newcommand{\compUnSeen}{\comp_\text{unseen}}
\newcommand{\attr}{\mathcal{A}}
\newcommand{\obj}{\mathcal{O}}
\newcommand{\sess}{\mathcal{S}}
\subsection{Problem Setup}
We begin by formalizing CZSL. Let $\attr$ and $\obj$ denote the attribute and object sets, with $\comp = \attr \times \obj = {(a,o)\mid a\in\attr, o\in\obj}$ as all possible compositions.
Let $\comp_\text{seen}, \comp_\text{unseen}\subset \comp$ be two disjoint subsets of $\comp$ consisting of \textit{seen} and \textit{unseen} compositions, $\comp_\text{seen} \cap \comp_\text{unseen} = \phi$.
Training set consists of images labeled with $\compSeen$  compositions; $\mathcal{T}_s= \{(x,c)|~x \in X,~c \in \compSeen\}$, 
testing set consists of images from both $\compSeen$ and $\compUnSeen$ compositions.
We follow \textit{closed-world} setup ~\cite{zhang2024continual} where testing label space is restricted to only $\comp_\text{test} = \compSeen \cup \compUnSeen$ avoiding ``invalid'' compositions.

\subsection{Continual-CZSL Setup.} 
In Continual-CZSL (CCZSL) the attributes, objects, and their composition are partitioned into $T+1$ sessions: $\sess^{(s)} = \left( \attr^{(s)}, \obj^{(s)}, \comp^{(s)} \right)$, where $s \in [0,T]$ and $\comp^{(s)}=\attr^{(s)} \times \obj^{(s)}$.  
We impose $\cup_{s=0}^T \attr^{(s)} = \attr$, $\cup_{s=0}^T \obj^{(s)} = \obj$, and $\cup_{s=0}^T \comp^{(s)} = \comp$.
For session $t$, the model is trained on $\mathcal{A}^{(t)}$, $\mathcal{O}^{(t)}$, and $\mathcal{C}^{(t)}_{\text{seen}}$. During training, it may access models from earlier sessions $0 \le s < t$.
Evaluation is performed on $\comp_\text{test}^{(t)} = \underset{0 \le s \le t}{\cup} {\compSeen^{(s)} \cup \compUnSeen^{(s)}}$, capturing current-session performance and prior knowledge retention.
We omit the session superscript when unambiguous.


\section{Proposed Framework}
\label{sec:proposed_framework}
Fig.~\ref{fig:framework} illustrates the proposed PromptCCZSL framework. 
Following~\cite{huang2024troika,li2024context}, 
our model leverages soft prompt tuning for primitives on a frozen VLM text encoder while adapting the vision backbone via parameter-efficient tuning.

\subsection{Primitive Concepts Learning.}
We maintain a session-shared \emph{Soft Prompt Bank} with one learnable embedding per primitive concept. 
To initialize each concept prompt, we tokenize its text using CLIP, discard special tokens (e.g., \texttt{[SOS]/[EOS]}), and use the mean of the remaining token embeddings as the initial soft prompt vector, which is subsequently fine-tuned. Collectively, these embeddings form dense attribute and object prompt banks:
$\mathbf{B}_{\text{a}} \in \mathbb{R}^{|\mathcal{A}|\times d}$ and 
$\mathbf{B}_{\text{o}} \in \mathbb{R}^{|\mathcal{O}|\times d}$. 
To model contextual semantics, each primitive embedding is concatenated with learnable prefix tokens 
$[b_1^a,\dots,b_m^a]$, 
$[b_1^o,\dots,b_m^o]$, and 
$[b_1^c,\dots,b_m^c]$, 
where $m$ denotes the context length. 
The resulting fully learnable prompts are:
$
\hat{b}_a^i = [b_1^a,\dots,b_m^a,\mathbf{B}_a[i]], 
\hat{b}_o^j = [b_1^o,\dots,b_m^o,\mathbf{B}_o[j]], 
\hat{b}_c^{ij} = [b_1^c,\dots,b_m^c,\mathbf{B}_a[i],\mathbf{B}_o[j]].
$

These prompts are mapped to textual embeddings via a pre-trained, frozen text encoder $f_{\text{text}}$ as $ \{e_a^i, e_o^j, e_c^{ij}\} = f_{\text{text}}(\hat{b}_a^i, \hat{b}_o^j, \hat{b}_c^{ij}). $
\subsection{Image Representation Learning. }
In parallel, the CLIP-pretrained vision encoder uses lightweight adapters to extract image features while preserving pretrained representations. 
Given an image $\mathbf{x}$, the encoder produces visual patch tokens 
$\mathbf{V} = [v_1,\dots,v_L] \in \mathbb{R}^{L\times d}$ 
and a global \texttt{[CLS]} token $v_{\mathrm{cls}} \in \mathbb{R}^d$. 
The composition embedding $v_c$ =  $v_{\mathrm{cls}}$, derived from the adapter-tuned \texttt{[CLS]} token, which captures joint attribute--object semantics.
A \emph{Primitive Decomposition Module} then decomposes this global representation into attribute, and object embeddings using two independent learnable affine projections $\mathbb{R}^{d}\!\to\!\mathbb{R}^{d}$. Formally, we compute $v_* = D_*(\mathbf{v_{\mathrm{cls}}})$ for $* \in \{a,o\}$ and set $v_c = v_{\mathrm{cls}}$. 
Each $D_*$ is a lightweight feed-forward head (FC + BN + ReLU + Dropout) followed by token pooling, producing a single vector in $\mathbb{R}^d$. These visual embeddings are then used to construct context-aware textual representations as follows.
\subsection{Session-Agnostic Fusion Module (SAgM2F)}
It refines attribute and object textual embeddings by applying cross-attention (CA) with visual patch tokens $(\textbf{V})$:
$
\tilde{e}_*^i = \mathrm{MLP}\big(\mathrm{CA}(e_*^i W_q, \mathbf{V} W_k, \mathbf{V} W_v)\big), * \in \{a,o\}.
$
where $W_q$, $W_k$, and $W_v$ are learnable projections for query, key, and value. Because primitives may recur across sessions, SAgM2F updates embeddings for all attributes and objects observed up to the current session, ensuring globally consistent refinement.

\subsection{Session-Aware Fusion Module (SAwM2F).}
Unlike attributes and objects, compositions do not repeat across sessions. 
Accordingly, composition embeddings are partitioned as
$
e_c = [\, e_{c,\text{head}} \,\Vert\, e_{c,\text{tail}} \,],
$
where $e_{c,\text{head}}$ contains compositions from previous sessions and 
$e_{c,\text{tail}}$ corresponds to newly introduced compositions in the current session.
Only the tail embeddings are refined via cross-attention with visual patches:
$
\hat{e}_{c,\text{tail}} = \mathrm{CA}(e_{c,\text{tail}} W_q, \mathbf{V} W_k, \mathbf{V} W_v),
$
followed by an MLP to obtain context-aware representations 
$\tilde{e}_{c,\text{tail}}$. 
The head embeddings are passed through an MLP without cross-attention.
Updating only the tail preserves prior semantics, enabling SAwM2F to learn new attribute–object compositions adaptively and maintain stable continual learning.
\subsection{Cross-Session Knowledge Distillation Loss (CSKD)}
We use CSKD to retain knowledge across sessions. During session $S^{(t)}$, model from $S^{(t-1)}$ is frozen and serves as a teacher to distill logits for attributes, objects, and compositions seen in previous sessions. This enforces compatibility with prior primitives while allowing the $S^{(t)}$ model as student to learn new attribute--object pairs.
Let \textit{overlapping primitives} be those shared between consecutive sessions; for attributes, this set is 
$\mathcal{A}^{(t-1)} \cap \mathcal{A}^{(t)}$. 
Denote the teacher and student logits over this set as $\mathbf{z}_a^{(t-1)}$ and $\mathbf{z}_a^{(t)}$, respectively. The cross-session distillation loss is defined as
\begin{equation}
\mathcal{L}^{(t)}_{\text{CSKD}}
=
\sum_{* \in \{a,o,c\}}
\tau^2 \, \mathrm{KL}\!\left(
\operatorname{softmax}\!\left(\mathbf{z}_*^{(t-1)}/\tau\right)
\,\|\, 
\operatorname{softmax}\!\left(\mathbf{z}_*^{(t)}/\tau\right)
\right).
\label{eq:CSKDloss}
\end{equation}
\subsection{Primitive-aware Anchor Alignment Loss (PAAL)}
PAAL enforces alignment between overlapping embeddings of attributes and objects and their anchors from the previous session, thereby stabilizing the prompt bank across sessions. Let $\mathbf{B}_a^{(t)}$ and $\mathbf{B}_o^{(t)}$ denote the attribute and object prompt banks in the current session, and $\mathbf{B}_a^{(t-1)}$ and $\mathbf{B}_o^{(t-1)}$ denote the corresponding teacher prompt banks. In Eq.~\ref{eq:PAALloss}, $\cos(\cdot,\cdot)$ represents the cosine similarity between normalized vectors. This loss preserves directional consistency of recurring attributes and objects:
\begin{equation}
\mathcal{L}^{(t)}_{\text{PAAL}} =
\sum_{* \in \{a,o\}} \sum_{i \in \mathcal{X}_*^{(t)} \cap \mathcal{X}_*^{(t-1)}} 
\big( 1 - \cos(\mathbf{B}_*^{(t)}[i], \mathbf{B}_*^{(t-1)}[i]) \big).
\label{eq:PAALloss}
\end{equation}
\subsection{Orthogonal Projection Loss (OPL)}
\begin{figure*}
\centering
\includegraphics[width=0.8\linewidth]{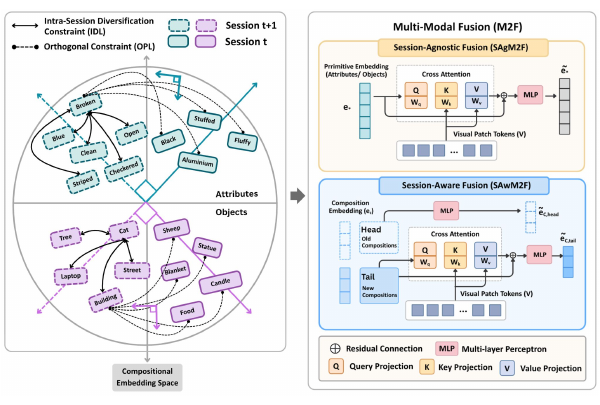}
\caption{\textbf{(left) Intra-session Diversification Loss (IDL) with Orthogonal Projection Loss (OPL).} IDL and OPL promote diverse, disentangled attribute and object embeddings across sessions. \textbf{(right) Multi-Modal Fusion (M2F) module.} SAgM2F refines primitive embeddings using cross-attention, while SAwM2F selectively updates new compositions to preserve prior knowledge.}
\label{fig:idl_opl_m2f}
\end{figure*}

PAAL stabilizes recurring primitives but does not prevent interference from new prompts. 
OPL enforces orthogonality between current and prior prompt subspaces, preserving compositional generalization (Fig.~\ref{fig:idl_opl_m2f}, left). At session $s$, the attribute prompt bank $\mathbf{B}_a$ contains all attributes observed so far. 
We partition it into previously learned prompts $\mathbf{B}_{a,\text{head}}$ and newly introduced prompts $\mathbf{B}_{a,\text{tail}}$ and defined loss in Eq. \ref{eq: opl} where $i$ and $j$ index the head and tail prompts and $\langle,\rangle$ represents the average cosine similarity between the two sets: 
\begin{equation}
\label{eq: opl}
\mathcal{L}^{(t)}_{\text{OPL}}
=
\sum_{* \in \{a,o\}}
\frac{1}{|\mathbf{B}_{*,\text{head}}|\,|\mathbf{B}_{*,\text{tail}}|}
\sum_i \sum_j 
|\left\langle 
\mathbf{B}_{*,\text{head}}[i], 
\mathbf{B}_{*,\text{tail}}[j]
\right\rangle|.
\end{equation}

\subsection{Intra-Session Diversification Loss (IDL)}
Similar to OPL which enforces independence between head and tail prompt bank, IDL encourages each new attribute and object to occupy a distinct semantic direction. Let $\mathbf{B}^{(s)}_{a\text{,tail}}$ and $\mathbf{B}^{(s)}_{o\text{,tail}}$ denote the subsets of attribute and object prompts introduced in the current session and defined loss in Eq. \ref{eq: idl}. By minimizing intra-session correlations, IDL promotes decorrelated and diverse prompt embeddings for newly added primitives, reducing representational overlap and enhancing compositional expressiveness.
\begin{equation}
\label{eq: idl}
\mathcal{L}^{(t)}_{\text{IDL}}
=
\sum_{* \in \{a,o\}}
\frac{1}{|\mathbf{B}^{(s)}_{*,\text{tail}}|^2 - |\mathbf{B}^{(s)}_{*,\text{tail}}|}
\sum_{i \neq j}
|\left\langle
\mathbf{B}^{(s)}_{*,\text{tail}}[i],
\mathbf{B}^{(s)}_{*,\text{tail}}[j]
\right\rangle|.
\end{equation}

\subsection{Training Flow}
Each continual session $\mathcal{S}^{(t)}$ expands the compositional space while retaining knowledge from prior sessions ${\mathcal{S}^{(0)},\dots,\mathcal{S}^{(t-1)}}$. The student at session $t$ is initialized from the frozen teacher of $\mathcal{S}^{(t-1)}$, ensuring continuity in the embedding space, and classifiers are expanded to include new compositions in $\mathcal{C}^{(t)}$. New primitives are regularized by the Orthogonal Projection Loss (OPL) and the Intra-Session Diversification Loss (IDL) to remain orthogonal to previous subspaces and diverse within the session. Newly introduced composition prompts are refined via SAwM2F, while attribute and object prompt banks are updated through SAgM2F to maintain global semantic consistency. Finally, three classification branches model conditional alignment
$P(\tilde e_* \mid v_*)$ for $* \in \{a,o,c\}$. 
Given visual embeddings $(v_a, v_o, v_c)$ and fused textual embeddings $(\tilde e_a^i, \tilde e_o^j, \tilde e_c^{ij})$, logits are computed using temperature-scaled cosine similarity:
$
z = \exp(\beta)\,\langle v, \tilde e \rangle,
$
where $\beta$ is the learned \texttt{logit\_scale}~\cite{radford2021learning}. 
For each branch, logits $\mathbf{z}_*$ over the corresponding label space 
($\mathcal A$, $\mathcal O$, $\mathcal C$) are normalized as
$
p(\tilde e_* \mid v_*) = \mathrm{softmax}(\mathbf{z}_*/\tau),
$
with temperature parameter $\tau$.
Each branch is optimized via cross-entropy with respect to the ground-truth label $y_*$ as
$
\mathcal{L}_* = 
-\frac{1}{|\mathcal{X}|} 
\sum_{x \in \mathcal{X}} 
\log p(y_* \mid v_*), 
* \in \{a,o,c\}.
$
The compositional objective is
\begin{equation}
\mathcal{L}_{\text{CE}} = \alpha_a \mathcal{L}_a + \alpha_o \mathcal{L}_o + \alpha_c \mathcal{L}_c.
\label{eq: CEloss}
\end{equation}
\subsubsection{Multi-Modal Fusion (M2F)}
Cross-modal alignment is achieved using attention-based fusion modules that refine textual embeddings with visual information. The model employs two mechanisms (Fig. \ref{fig:idl_opl_m2f}, right):
\textbf{Session-Agnostic M2F (SAgM2F):}
For the attribute and object branches, SAgM2F enables each textual embedding to attend to visual patch tokens in a session-independent manner: $
\tilde{e}_a^i = \text{CA}(e_a^i, V), \quad
\tilde{e}_o^j = \text{CA}(e_o^j, V),
$ where $\text{CA}(\cdot)$ denotes a scaled dot-product cross-attention operation followed by a lightweight MLP:
\begin{equation}
\label{eq:CA_open}
\text{CA}(e^l, V) =
\mathrm{softmax}\!\left(\frac{(e^l W_q)(V W_k)^\top}{\sqrt{d}}\right)(V W_v),
\end{equation}
with $V$ denoting the set of visual patch tokens without [CLS] token, and $W_q, W_k, W_v \in \mathbb{R}^{d\times d}$ being learnable projection matrices.  
Here, the textual embedding $e^l$ acts as the query, while the visual features serve as keys and values.  
This mechanism ensures that attribute and object representations remain globally consistent across sessions.

\textbf{Session-Aware M2F (SAwM2F):}
For the composition branch, the current session contains both previously seen (\textit{head}) and newly introduced (\textit{tail}) attribute--object pairs.  
New pairs often exhibit session-specific visual variations, so SAwM2F refines only the tail embeddings to adapt to these changes while preserving stability for the head prompts.  
Following the cross-attention in Eq.~\ref{eq:CA_open}, the refined tail and frozen head embeddings are concatenated and projected through a lightweight MLP:
$
\tilde{e}_c = \mathrm{MLP}\big([\hat{e}_{c,\text{tail}}, \hat{e}_{c,\text{head}}]\big).
$
By refining only tail embeddings and keeping the head fixed, SAwM2F learns new attribute--object compositions while preserving prior semantics and stable continual learning.


%

\begin{table}[h]
\centering
\caption{Session splits of UT-Zappos, C-GQA, and MIT-States. Values are reported in session order.}
\label{tab:Table_1}
\setlength{\tabcolsep}{3pt}
\renewcommand{\arraystretch}{0.85}
\begin{tabular}{lccc}
\toprule
 & \textbf{UT-Zappos} & \textbf{C-GQA} & \textbf{MIT-States} \\
\midrule
Session & 0,1,2 & 0,1,2,3,4,5 & 0,1,2,3 \\
Attr    & 8,4,4 & 233,35,32,36,39,38 & 37,25,26,27 \\
Obj     & 6,3,3 & 363,58,67,64,62,60 & 71,61,58,55 \\
Train   & 24,27,32 & 2392,491,772,836,562,539 & 161,278,361,462 \\
Val     & 7,10,13 & 958,168,358,366,225,217 & 76,145,164,215 \\
Test    & 9,14,13 & 730,172,265,275,166,203 & 90,187,229,294 \\
\bottomrule
\end{tabular}
\end{table}

\subsubsection{Overall Training Objective:}
\noindent The overall training objective
is defined in Eq. \ref{eq:total}, where $\lambda_{\text{ce}}$,$\lambda_{\text{kd}}$, $\lambda_{\text{PAAL}}$, $\lambda_{\text{opl}}$, and $\lambda_{\text{idl}}$ are scalar weights controlling the contribution of each term:
\begin{equation}
\mathcal{L}_{\text{total}} =
\lambda_{\text{ce}} \mathcal{L}_{\text{CE}}
+ \lambda_{\text{kd}} \mathcal{L}_{\text{CSKD}}
+ \lambda_{\text{PAAL}} \mathcal{L}_{\text{PAAL}}
+ \lambda_{\text{opl}} \mathcal{L}_{\text{OPL}}
+ \lambda_{\text{idl}} \mathcal{L}_{\text{IDL}}.
\label{eq:total}
\end{equation}

\subsubsection{Inference}
At test time, each image is evaluated against the union of all seen
$\bigcup_{k=0}^{T}\mathcal{C}_k$ and unseen compositions, activating all composition embeddings learned across sessions. Final predictions are obtained by scoring each image against every composition as
\begin{equation}
\hat{c} = 
\arg\max_{c \in \mathcal{C}}
\Big(
\lambda_{\text{c}} \, p(\tilde{e}_c | v_c) +
\lambda_{\text{a}} \, p(\tilde{e}_a | v_a) \cdot
\lambda_{\text{o}} \, p(\tilde{e}_o | v_o)
\Big)
\label{eq:inference}
\end{equation}
where $\lambda_c$, $\lambda_a$, and $\lambda_o$ are weighting coefficients.
\subsection{Datasets}
\label{subsec}

We evaluate our method on three compositional zero-shot learning benchmarks: MIT-States~\cite{isola2015discovering}, UT-Zappos~\cite{yu2014fine}, and C-GQA~\cite{naeem2021learning}, following the continual compositional zero-shot learning (CCZSL) setting~\cite{zhang2024continual}. For UT-Zappos and C-GQA, we adopt the splits and evaluation protocol of the CCZSL benchmark~\cite{zhang2024continual}. For MIT-States, we use the standard CZSL splits with standardized criteria consistent with prior work.
UT-Zappos contains 50{,}025 images across 16 attributes and 12 objects, while C-GQA contains 38{,}000 images with 413 attributes and 674 objects. MIT-States covers diverse objects and state attributes and is evaluated using standardized splits to mitigate annotation noise and dataset bias. Tab.~\ref{tab:Table_1} summarizes the session-wise splits. Due to computational constraints, C-GQA results are reported up to session $S^{(4)}$.

\subsection{Evaluation Metrics}
\label{subsec:evaluation_metrics}
We follow the standard CZSL evaluation protocol~\cite{naeem2021learning, purushwalkam2019task, zhang2022learning, chao2016empirical} in the continual setting, evaluating adaptation to new compositions while retaining previously acquired knowledge. At session $t$, the model is evaluated on accumulated test data from sessions $0$ to $t$. We report top-1 best-seen ($S$), best-unseen ($U$), harmonic mean ($H$), attribute (\textit{Attr}), and object (\textit{Obj}) accuracies, along with session-wise and average $AUC$ for continual performance.

\subsubsection{Catastrophic Forgetting in CCZSL (CFZSEval).}
We measure catastrophic forgetting on unseen compositions by evaluating if the model retains correct predictions from earlier sessions.
Following \cite{kang2025advancing}, let $\Xi_i$ be test-samples labeled with unseen (zero-shot) compositions in session $i$ and $M_t$ the model trained on session $t$. Denoting $\text{AUC}_i^t$ as the AUC computed on $\Xi_i$ using $M_t$ for $0 \le i \le t$, the average forgetting is defined as
\begin{equation}
\label{eq:avg_fau}
\mathcal{F}_{\text{AUC}}^i = \frac{1}{T-i-1} \sum_{t=i+1}^{T-1} \Big| \text{AUC}_i^i - \text{AUC}_i^t \Big| \,, \quad
\mathcal{F}_{\text{AUC}} = \frac{1}{T-1} \sum_{i=0}^{T-2} \mathcal{F}_{\text{AUC}}^i.
\end{equation}
Larger $\mathcal{F}_{AUC}$ means more forgetting, therefore lower is better. Note that, objects and attributes might reoccur in next sessions, that's why we evaluate catastrophic forgetting using $\mathcal{F}_{AUC}^{i}$ where $i^{th}$ represents the $S^{(i)}$ session data (Tab.~\ref{tab:Table_6},~\ref{tab:Table_7}, ~\ref{tab:Table_8}). 
\subsection{Baselines and Compared Methods}
\label{subsubsec:implementation_details_baselines}
We use Zhang et al.~\cite{zhang2024continual} as the non-VLM continual CZSL baseline. We also adapt~\cite{huang2024troika} as a VLM-based distillation baseline (\textit{CCZSL-Baseline}) and report an \textit{Oracle-CCZSL} trained on all data up to session $t$.
\begin{table*}[t]
\centering
\caption{Comparison with SOTA non-VLM CZSL methods in continual setting (results are reported from \cite{zhang2024continual}). Session-wise AUC $\uparrow$, avg. AUC $\uparrow$, and Final gain $\uparrow$ of our method over prior methods are reported. $\dagger$ represents ours implementation (details in Sec. \ref{subsec:sota_results}). Our method, PromptCCZSL, outperforms both VLM-based and non-VLM-based methods, with substantial performance gains across all benchmarks.}
\label{tab:Table_2}
\resizebox{1.0\textwidth}{!}{
\begin{tabular}{ll|ccccc|ccccccc}
\toprule
\multicolumn{2}{c|}{} &
\multicolumn{5}{c|}{UT-Zappos~\cite{yu2014fine}} &
\multicolumn{7}{c}{C-GQA~\cite{naeem2021learning}} \\
\cmidrule(lr){3-7}
\cmidrule(lr){8-14}
Method  & Data &
$S^{(0)}$ & $S^{(01)}$ & $S^{(012)}$ &  &  &
$S^{(0)}$ & $S^{(01)}$ & $S^{(012)}$ & $S^{(0123)}$ & $S^{(01234)}$  &  &  \\
& Model &
$S^{(0)}$ & $S^{(1)}$ & $S^{(2)}$ & Avg & Final &
$S^{(0)}$ & $S^{(1)}$ & $S^{(2)}$ & $S^{(3)}$ & $S^{(4)}$  & Avg & Final  \\
\midrule
\multicolumn{14}{c}{Non-VLM based Methods} \\
\midrule
CCZSL--AoP        & ECCV’18   & 42.21 & 20.34 & 16.25 & 26.27 & +14.25  &  2.52 &  1.32 & 0.97 & 0.50 & 0.34 & 1.13 & +12.08 \\
CCZSL--SymNet     & CVPR’20   & 41.93 & 13.58 &  9.46 & 21.66 & +18.88  &  3.42 &  2.28 & 1.15 & 0.65 & 0.63 & 1.63 & +11.58 \\
CCZSL--VisProdNN  & NeurIPS’21& 43.59 & 17.65 &  2.78 & 21.34 & +19.18  &  4.18 &  0.40 & 1.19 & 0.44 & 0.24 &  1.29 & +11.92 \\
CCZSL--SCEN       & CVPR’22   & 44.00 & 15.61 &  8.74 & 22.78 & +17.74  &  3.43 &  0.64 & 0.75 & 0.26 & 0.11 & 1.04 & +12.17 \\
CCZSL--CANet      & CVPR’23   & 45.48 & 19.89 &  5.05 & 23.47 & +17.05  &  5.17 &  2.49 & 1.90 & 1.17 & 1.27 &  2.4 & +10.81 \\
Zhang~\textit{et al.} & IJCAI’24 & 47.70 & 24.73 & 18.96 & 30.46 & +10.46  &  5.07 &  3.89 & 3.88 & 2.74 & 2.32 & 3.58 & +9.63 \\
\midrule
\multicolumn{14}{c}{VLM based Methods} \\
\midrule
Clip-Zeroshot$\dagger$  & CVPR’24 & 29.61 & 4.59 &  4.9 & 13.03 & +27.49  &  2.21 & 2.25  & 1.74 & 1.47 & 1.51 & 1.84 & +11.37 \\

CCZSL-CSP$\dagger$  & CVPR’24  & 48.46 & 19.19 & 3.27 & 23.64  & +16.88  &  10.39 &  5.16 & 4.14 & 2.09 & 1.71 & 4.70 & +8.51 \\

CCZSL-DFSP$\dagger$  & CVPR’24  & 44.39 & 21.57 & 1.47 & 22.47 & +18.04  & 14.99  & 6.75  & 4.08 & 1.88 & 1.24 & 5.79 & +7.42  \\

CCZSL-Baseline$\dagger$&   & 60.42 & 15.39 & 1.61 & 25.81  & +14.71 & 16.58 & 2.16 & 3.82 & 2.68 & 2.75 & 5.60 & +7.61 \\
\textbf{Prompt-CCZSL (ours)}&  &
\textbf{60.42} & \textbf{40.10} & \textbf{21.06} & \textbf{40.52} & -- &
\textbf{16.58} & \textbf{14.53} & \textbf{13.66} & \textbf{11.44} & \textbf{9.83} & \textbf{13.21} & -- \\
\bottomrule
\end{tabular}
}
\end{table*}

\begin{table*}[t]
\centering
\caption{CCZSL results on UT-Zappos~\cite{yu2014fine}. Cumulative performance is evaluated on the test set of zero-shot compositions up to session $S^{(t)}$. We report AUC, best Seen (S), Unseen (U), and Harmonic Mean (HM). PromptCCZSL consistently outperforms the baseline and state-of-the-art methods, achieving performance closer to the oracle.}
\label{tab:Table_3}
\resizebox{0.9\textwidth}{!}{
\begin{tabular}{l|cccc|cccc|cccc}
\toprule
\multirow{1}{*}{\textbf{Method}} &
\multicolumn{4}{c|}{\textbf{$S^{(0)}$ Model / $S^{(0)}$ Data}} &
\multicolumn{4}{c|}{\textbf{$S^{(1)}$ Model / $S^{(01)}$ Data}} &
\multicolumn{4}{c}{\textbf{$S^{(2)}$ Model / $S^{(012)}$ Data}} \\
& AUC $\uparrow$& S $\uparrow$& U $\uparrow$& HM $\uparrow$ 
& AUC $\uparrow$& S $\uparrow$& U $\uparrow$& HM $\uparrow$ 
& AUC $\uparrow$& S $\uparrow$& U $\uparrow$& HM $\uparrow$  \\ 
\midrule

Clip-Zeroshot 
& 29.61 & 50.0	& 72.01 & 44.78 
& 4.59	& 11.07	& 55.14 & 14.46
& 4.9	& 15.54	& 49.02 & 15.33\\

CCZSL-CSP 
& 48.46 & 78.76 & 80.6 & 57.69 
& 19.19 & 48.41 & 48.65 & 36.38 
& 3.27 & 15.05 & 31.78 & 14.05\\

CCZSL-DFSP
& 44.39	& 77.69 & 75.47 & 55.42
& 21.57	& 47.82	& 51.67 & 40.41
& 1.47	& 10.36	& 22.79 & 9.33
\\

CCZSL--Baseline 
& 60.42 & 83.33 & 81.62 & 68.50
& 15.39 & 48.41 & 44.15 & 31.11
& 1.61 & 0.64 & 29.67 & 0.87 \\

\textbf{Prompt-CCZSL (ours)} 
& 60.42 & 83.33 & 81.62 & 68.50
& \textbf{40.10} & \textbf{73.97} & \textbf{64.14} & \textbf{52.56}
& \textbf{21.06} & \textbf{38.42} & \textbf{68.59} & \textbf{37.33} \\
\midrule

CCZSL--Oracle
& 60.42 & 83.33 & 81.62 & 68.50
& 44.22 & 76.8 & 65.3 & 57.69
& 38.19 & 65.49 & 72.03 & 52.04
\\
\bottomrule
\end{tabular}}
\end{table*}
\begin{table*}[t]
\centering
\caption{CCZSL results on MIT-States~\cite{isola2015discovering}. Cumulative performance is evaluated on the test set of zero-shot compositions up to session $S^{(t)}$. We report AUC, best Seen (S), Unseen (U), and Harmonic Mean (HM). PromptCCZSL consistently outperforms the baseline and state-of-the-art methods, achieving performance closer to the oracle.}
\label{tab:Table_4}
\resizebox{\textwidth}{!}{
\begin{tabular}{l|cccc|cccc|cccc|cccc}
\toprule

\multirow{1}{*}{\textbf{Method}} &
\multicolumn{4}{c|}{\textbf{$S^{(0)}$ Model / $S^{(0)}$ Data}} &
\multicolumn{4}{c|}{\textbf{$S^{(1)}$ Model / $S^{(01)}$ Data}} &
\multicolumn{4}{c|}{\textbf{$S^{(2)}$ Model / $S^{(012)}$ Data}} &
\multicolumn{4}{c}{\textbf{$S^{(3)}$ Model / $S^{(0123)}$ Data}} \\

& AUC $\uparrow$& S $\uparrow$& U $\uparrow$& HM $\uparrow$ 
& AUC $\uparrow$& S $\uparrow$& U $\uparrow$& HM $\uparrow$ 
& AUC $\uparrow$& S $\uparrow$& U $\uparrow$& HM $\uparrow$ 
& AUC $\uparrow$& S $\uparrow$& U $\uparrow$& HM $\uparrow$  \\
\midrule

Clip-Zeroshot
&  29.03 & 49.07 & 69.05 & 45.96
&  15.94 & 36.8	 & 54.94 & 32.36
&  17.15 & 38.47 & 53.42 & 33.84
&  11.01 & 30.13 & 46.01 & 26.06 \\

CCZSL-CSP
& 37.34	& 60.97	& 69.24 & 52.51
& 11.59	& 32.71	& 48.26 & 27.64
& 18.6	& 45.12	& 49.31 & 35.39
& 11.92	& 36.13	& 41.97 & 27.7 \\

CCZSL-DFSP
& 38.33	& 62.45	& 69.9 & 52.55
& 9.43	& 27.88	& 48.64 & 22.86
& 12.46	& 36.25	& 43.76 & 28.01 
& 7.93	& 30.04	& 33.08 & 22.78\\

CCZSL-Baseline
& 44.41 & 68.03 & 73.19 & 57.96
& 26.49 & 53.28 & 58.11 & 42.72
& 12.62 & 36.99 & 41.41 & 29.16
& 7.74  & 29.87 & 32.65 & 22.79 \\

\textbf{Prompt-CCZSL (ours)}
& 44.41 & 68.03 & 73.19 & 57.96
& \textbf{32.38} &  \textbf{59.59} &  \textbf{63.01} &  \textbf{47.01}
& \textbf{24.05} &  \textbf{50.24} &  \textbf{55.23} &  \textbf{41.12}
& \textbf{18.50} &  \textbf{45.17} &  \textbf{49.67} &  \textbf{35.23} \\
\midrule
CCZSL--Oracle
& 44.41 & 68.03 & 73.19 & 57.96
& 34.23 & 60.44 & 64.33 & 49.49
& 27.33 & 54.61 & 57.32 & 43.95
& 21.3  & 49.79 & 50.16 & 38.6 \\
\bottomrule
\end{tabular}}
\end{table*}
\begin{table*}[t]
\centering
\caption{CCZSL results on C-GQA~\cite{naeem2021learning}. Cumulative performance is evaluated on zero-shot compositions up to session $S^{(t)}$ using AUC, Seen (S), Unseen (U), and Harmonic Mean (HM). PromptCCZSL consistently outperforms prior methods and approaches oracle performance.}
\label{tab:Table_5}
\resizebox{\textwidth}{!}{
\begin{tabular}{l|
cccc|cccc|cccc|cccc|cccc}
\toprule
\multirow{2}{*}{\textbf{Method}} &
\multicolumn{4}{c|}{\textbf{$S^{(0)}$ Model / $S^{(0)}$ Data}} &
\multicolumn{4}{c|}{\textbf{$S^{(1)}$ Model / $S^{(01)}$ Data}} &
\multicolumn{4}{c|}{\textbf{$S^{(2)}$ Model / $S^{(012)}$ Data}} &
\multicolumn{4}{c|}{\textbf{$S^{(3)}$ Model / $S^{(0123)}$ Data}} &
\multicolumn{4}{c}{\textbf{$S^{(4)}$ Model / $S^{(01234)}$ Data}}\\

& AUC $\uparrow$& S $\uparrow$& U $\uparrow$& HM $\uparrow$ 
& AUC $\uparrow$& S $\uparrow$& U $\uparrow$& HM $\uparrow$ 
& AUC $\uparrow$& S $\uparrow$& U $\uparrow$& HM $\uparrow$ 
& AUC $\uparrow$& S $\uparrow$& U $\uparrow$& HM $\uparrow$ 
& AUC $\uparrow$& S $\uparrow$& U $\uparrow$& HM $\uparrow$ \\
\midrule

CLIP-Zeroshot  
& 2.21 & 10.33 & 30.21 & 10.62
& 2.25 & 10.15 & 30.29 & 10.95
& 1.74 & 8.5 & 27.75 & 9.5
& 1.47 & 7.36 & 26.74 & 8.72
& 1.51 & 7.59 & 26.71 & 8.85
\\

CCZSL-CSP
& 10.39 &	38.01 &	34.58 & 26.26
& 5.16  &	23.04 & 29.17 & 18.35
& 4.14  &   22.34 &	23.43 & 16.43
& 2.09  &	15.28 &	19.24 & 11.44 
& 1.71	&   12.33 &	18.4  & 10.34 
\\
CCZSL-DFSP
& 14.99	& 45.71	& 39.17 & 32.21
& 6.75	& 26.87	& 32.21 & 20.98
& 4.08	& 23.2	& 20.81 & 17.0
& 1.88	& 17.08	& 13.59 & 10.93
& 1.24	& 13.14	& 11.47 & 9.1
\\
CCZSL--Baseline &
16.58  & 45.37 & 42.71 & 34.57 & 
2.16  & 9.67 & 27.08 & 11.19 &
3.82  & 23.39 & 19.24 & 16.51 &
2.68  & 20.45 & 16.2 & 13.41 &
2.75  & 19.17 & 17.41 & 13.81 \\

\textbf{Prompt-CCZSL (ours)} &
16.58  & 45.37 & 42.71 & 34.57 &
\textbf{14.53} & \textbf{41.79} & \textbf{40.71} & \textbf{32.02} & 
\textbf{13.66} & \textbf{43.39} & \textbf{37.3} & \textbf{30.94} & 
\textbf{11.44} & \textbf{37.09} & \textbf{36.63} & \textbf{28.09} & 
\textbf{9.83}  & \textbf{33.28} & \textbf{35.31} & \textbf{25.98} 
\\
\midrule
Oracle-CCZSL &
16.58 & 45.37 & 42.71 & 34.57 &
15.56 & 43.23 & 41.82 & 33.28 &
14.96 & 45.77 & 37.83 & 31.96 &
13.96 & 42.02 & 38.59 & 31.47 &
13.52 & 41.55 & 38.83 & 30.74\\
\bottomrule
\end{tabular}}
\end{table*}
We compare our approach with several representative CZSL and CCZSL baselines. The comparison methods are categorized into non-VLM-based and VLM-based approaches. For non-VLM-based methods, we report the CCZSL results reported by Zhang et al.~\cite{zhang2024continual}, who extended several classical CZSL methods to the continual compositional zero-shot learning setting. Since the implementation details and source code associated with these CCZSL adaptations are not publicly available, direct reproduction of these methods was not feasible. Therefore, we use directly the reported results from Zhang et al.~\cite{zhang2024continual} as reference baselines. In contrast, we implement and evaluate VLM-based CCZSL baselines using the same dataset splits adopted in this work. To promote reproducible research, we will publicly release the dataset splits and source code for all experiments on UT-Zappos, C-GQA, and MIT-States.

The comparison methods are as follows: (1) \textbf{AOP}~\cite{nagarajan2018attributes} models attributes as transformation operators applied to objects. (2) \textbf{SymNet}~\cite{li2020symmetry} introduces symmetry constraints for attribute--object transformations. (3) \textbf{VisProdNN}~\cite{karthik2021revisiting} formulates CZSL as a multi-task problem by predicting attributes and objects separately. (4) \textbf{SCEN}~\cite{Li_2022_CVPR} learns attribute--object embeddings in a contrastive space and enhances compositional diversity through a State Transition Module. (5) \textbf{CANet}~\cite{Wang_2023_CVPR} learns object-conditioned attribute representations using an attribute hyper-network. (6) \textbf{Zhang et al.}~\cite{zhang2024continual} address continual compositional zero-shot learning through learnable Super-Primitives for contextual adaptation and dual knowledge distillation to alleviate catastrophic forgetting.
We further compare with VLM-based approaches. (7) \textbf{CLIP Zero-Shot}~\cite{radford2021learning} directly leverages pretrained CLIP for compositional recognition through text prompts. 
Since no VLM-based baselines have been available specifically for continual compositional zero-shot learning (CCZSL), we extend representative state-of-the-art CZSL methods to the continual setting by incorporating knowledge distillation, a widely adopted strategy for mitigating catastrophic forgetting in continual learning~\cite{li2017learning,rebuffi2017icarl}. 
Specifically, (8) \textbf{CCZSL-CSP} extends CSP~\cite{nayaklearning} to the CCZSL setting, learning compositional prompt tokens for attributes and objects. (9) \textbf{CCZSL-DFSP} extends DFSP~\cite{lu2023decomposed}, which employs decomposed cross-modal fusion to improve image-text alignment. Both CCZSL-CSP and CCZSL-DFSP are evaluated under the CCZSL setting, providing strong VLM-based baselines for comparison with our \textbf{Prompt-CCZSL}.

\subsection{Implementation Details.}
Following~\cite{huang2024troika}, we implement PromptCCZSL in PyTorch with a frozen CLIP ViT-L/14 backbone~\cite{paszke2019pytorch}, optimizing only lightweight components such as visual prompts and vision-side adapters. We evaluate on UT-Zappos~\cite{yu2014fine}, C-GQA~\cite{naeem2021learning}, and MIT-States~\cite{isola2015discovering}.
Each session is trained for 15 epochs on an NVIDIA V100 32GB GPU using Adam with a learning rate of $5\times10^{-4}$, weight decay of $1\times10^{-5}$, and StepLR with step size 5 and decay 0.5. Batch sizes are 16 for UT-Zappos and 4 for C-GQA and MIT-States, except C-GQA session 4, where it is reduced to 2. Prompt context length is set to 8.
Loss weights are set to 0.65, 0.35, 0.05, 0.05, and 0.005 for knowledge distillation, cross-entropy, PAAL, OPL, and IDL, respectively. The first session is trained from scratch, with subsequent sessions initialized from the previous one.

\subsection{Results}
\label{subsec:results_section}
\subsubsection{Comparison with State-of-the-Art Methods.}
\label{subsec:sota_results}
Tab.~\ref{tab:Table_2} reports AUC comparison with prior non-VLM and adopted VLM continual CZSL approaches on UT-Zappos and C-GQA. Prompt-CCZSL consistently outperforms all baselines across every continual session, achieving highest average AUC on both benchmarks. Compared with Zhang et al.~\cite{zhang2024continual}, our method yields substantial improvements of +10.46\% AUC on UT-Zappos (30.46\% $\rightarrow$ 40.52\%) and +9.63\% AUC on C-GQA (3.58\% $\rightarrow$ 13.21\%). Notably, relative improvement on C-GQA highlighting robustness of PromptCCZSL under challenging large-scale compositional settings. While VLM based methods already demonstrate stronger generalization than earlier non-VLM methods~\cite{nagarajan2018attributes,li2020symmetry,karthik2021revisiting,li2022siamese,wang2023learning}, our prompt-based adaptation further amplifies this advantage. 
Compared with existing VLM-based approaches, including Clip-Zero Shot, CCZSL-CSP, and CCZSL-DFSP, the gains are consistent across all continual stages, indicating improved stability–plasticity~\cite{jung2023new} balance rather than isolated improvements at final session. 

\subsubsection{Comparison with CCZSL and Oracle Baselines.}

Tab.~\ref{tab:Table_3}
 --Tab. \ref{tab:Table_5} present the cumulative performance of VLM-based methods on UT-Zappos, C-GQA, and MIT-States respectively, benchmarking against baselines and oracle models in closed-world setting. Prompt-CCZSL consistently achieves the strongest performance among all continual CZSL baselines across different sessions and datasets, demonstrating its effectiveness in preserving previously acquired compositional knowledge while adapting to newly introduced compositions.
On UT-Zappos (Tab.~\ref{tab:Table_3}), Prompt-CCZSL attains the highest AUC among all continual methods after the initial session. Specifically, at $S^{(1)}$, it improves the AUC from 21.57\% achieved by CCZSL-DFSP to 40.10\%, while simultaneously increasing the harmonic mean (HM) from 40.41\% $\rightarrow$ 52.56\%. The advantage remains substantial at the most challenging session $S^{(2)}$, where Prompt-CCZSL obtains 21.06\% AUC and 37.33\% HM, outperforming the strongest baseline by large margins.

On MIT-States (Tab. \ref{tab:Table_4}), Prompt-CCZSL also shows the highest performance. Compared with strongest baseline (CCZSL-Baseline), the proposed method improves AUC; 26.49\% $\rightarrow$ 32.38\% at $S^{(1)}$, 12.62\% $\rightarrow$ 24.05\% at $S^{(2)}$, and 7.74\% $\rightarrow$ 18.50\% at $S^{(3)}$. The corresponding HM gains are also significant, increasing 42.72\% $\rightarrow$ 47.01\%, 29.16\% $\rightarrow$ 41.12\%, and 22.79\% $\rightarrow$ 35.23\%, respectively. 

A similar trend is observed on the more challenging C-GQA benchmark (Tab. \ref{tab:Table_5}). While existing continual CZSL methods experience severe performance degradation as the number of sessions increases, Prompt-CCZSL maintains substantially higher accuracy across all sessions. At $S^{(1)}$, the proposed method achieves 14.53\% AUC compared with 6.75\% for the CCZSL-DFSP, corresponding to a higher gain. The method continues to outperform others in later sessions with high margins, achieving AUC of 13.66\%, 11.44\%, and 9.83\% at $S^{(2)}$, $S^{(3)}$ and $S^{(4)}$ respectively. The consistently higher HM scores further suggest a better balance between seen and unseen composition recognition, highlights the robustness of the proposed framework under large-scale compositional shifts.
Consequently, Prompt-CCZSL narrows the gap to the Oracle upper bound and establishes a new state of the art for continual compositional zero-shot learning under the closed-world setting.

\subsubsection{Catastrophic Forgetting Analysis}
\label{subsubsec:cfg-analysis}
Tab.~\ref{tab:Table_6}--Tab.~\ref{tab:Table_8} compare catastrophic forgetting on UT-Zappos, MIT-States, and C-GQA. Across all datasets, CCZSL-Baseline rapidly forgets earlier compositions, whereas PromptCCZSL preserves higher AUC and attribute/object accuracy with lower $\mathcal{F}_{AUC}$.
On UT-Zappos, after learning $S^{(1)}$, the baseline AUC on $S^{(0)}$ drops to $0.0$, yielding $\mathcal{F}_{AUC}=60.42\%$, while PromptCCZSL retains $37.35\%$ AUC and reduces forgetting to $38.04\%$. After $S^{(2)}$, forgetting on $S^{(1)}$ decreases from $52.98\%$ to $4.22\%$.
A similar trend appears on MIT-States, where PromptCCZSL reduces forgetting for $S^{(0)}$, $S^{(1)}$, and $S^{(2)}$ from $41.81\%$, $37.39\%$, and $34.84\%$ to $29.72\%$, $17.85\%$, and $11.81\%$, respectively. On C-GQA, forgetting is also substantially reduced, including $16.34\%\rightarrow5.04\%$ for $S^{(0)}$, $14.68\%\rightarrow6.07\%$ for $S^{(1)}$, and $16.65\%\rightarrow2.14\%$ for $S^{(2)}$.

Overall, PromptCCZSL provides a better stability--plasticity trade-off by retaining earlier compositional knowledge while adapting to new attribute--object pairs. Fig.~\ref{fig:utzappos_qres} provides corresponding qualitative examples.
\begin{table*}[t]
\centering
\caption{Catastrophic forgetting comparison on UT-Zappos. Each model is evaluated on its session-specific test set. CCZSL-Baseline shows severe forgetting, while PromptCCZSL better preserves prior knowledge, achieving higher AUC and Attr/Obj accuracies with lower forgetting ($\mathcal{F}_{AUC}$, Eq.~\ref{eq:avg_fau}). Under cumulative zero-shot evaluation up to $S^{(t)}$, PromptCCZSL consistently outperforms the baseline.} 
\label{tab:Table_6}
\resizebox{1.0\textwidth}{!}{
\begin{tabular}{c|cccc|cccc|cccc|c}
\toprule
\multicolumn{14}{c}{\textbf{CCZSL -- Baseline}} \\
\midrule
\textbf{Data\textbackslash\ Model} 
& \multicolumn{4}{c|}{\textbf{Model $\mathcal{S}^{(0)}$}}
& \multicolumn{4}{c|}{\textbf{Model $\mathcal{S}^{(1)}$}}
& \multicolumn{4}{c|}{\textbf{Model $\mathcal{S}^{(2)}$}} 
& $\mathcal{F}_{AUC}^{i}\downarrow$\\
& AUC $\uparrow$ & Attr $\uparrow$& Obj $\uparrow$& HM $\uparrow$
& AUC $\uparrow$ & Attr $\uparrow$& Obj $\uparrow$& HM $\uparrow$
& AUC $\uparrow$ & Attr $\uparrow$& Obj $\uparrow$& HM $\uparrow$
&\\
\midrule
\textbf{Data $\mathcal{S}^{(0)}$} 
& 60.42 & 48.55 & 86.91 & 68.50
& 0.0 & 46.12 & 17.04 & 0.0
& 0.0 & 36.84 & 26.11 & 0.0
& \textcolor{red}{60.42}\\
\textbf{Data $\mathcal{S}^{(1)}$}
& -- & -- & -- & --
& 52.98 & 60.15 & 82.73 & 59.24
& 0.0 & 39.96 & 11.34 & 1.19
& \textcolor{red}{52.98}\\
\textbf{Data $\mathcal{S}^{(2)}$}
& -- & -- & -- & --
& -- & -- & -- & --
& 32.07 & 34.58 & 7.8 & 46.83
& --\\
\midrule

\multicolumn{13}{r}{} & $\mathcal{F}_{AUC}=$\textcolor{red}{56.7}\\
\midrule
& \multicolumn{4}{c|}{\textbf{Model $S^{(0)}$ / Data $S^{(0)}$}}
& \multicolumn{4}{c|}{\textbf{Model $S^{(1)}$ / Data $S^{(01)}$}}
& \multicolumn{4}{c|}{\textbf{Model $S^{(2)}$ / Data $S^{(012)}$}}\\
\textbf{Cumulative Data}
& 60.42 & 48.55 & 86.91 & 68.50
& 15.39 & 49.02 & 43.28 & 31.11
& 1.61 & 37.47 & 30.30 & 0.87
& \\
\toprule
\multicolumn{14}{c}{\textbf{PromptCCZSL (ours)}} \\
\midrule
\textbf{Data\textbackslash\ Model} 
& \multicolumn{4}{c|}{\textbf{Model $\mathcal{S}^{(0)}$}}
& \multicolumn{4}{c|}{\textbf{Model $\mathcal{S}^{(1)}$}}
& \multicolumn{4}{c|}{\textbf{Model $\mathcal{S}^{(2)}$}} 
& $\mathcal{F}_{AUC}^{i}\downarrow$\\
& AUC $\uparrow$ & Attr $\uparrow$& Obj $\uparrow$& HM $\uparrow$
& AUC $\uparrow$ & Attr $\uparrow$& Obj $\uparrow$& HM $\uparrow$
& AUC $\uparrow$ & Attr $\uparrow$& Obj $\uparrow$& HM $\uparrow$
&\\
\midrule
\textbf{Data $\mathcal{S}^{(0)}$}
& 60.42 & 48.55 & 86.91 & 68.50
& 37.35 & 55.61 & 66.9 & 52.39
& 7.41 & 45.64 & 47.44 & 18.97
& \textcolor{red}{\textbf{38.04}}\\
\textbf{Data $\mathcal{S}^{(1)}$}
& -- & -- & -- & --
& 34.82 & 57.34 & 77.42 & 47.06
& 30.6 & 58.06 & 56.71 & 46.41
& \textcolor{red}{\textbf{4.22}}\\
\textbf{Data $\mathcal{S}^{(2)}$}
& -- & -- & -- & --
& -- & -- & -- & --
& 13.8 & 33.2 & 75.44 & 28.19
& --\\
\midrule
\multicolumn{13}{r}{} 
& $\mathcal{F}_{AUC} =$\textcolor{red}{\textbf{21.13} ($-35.57$)}\\
\midrule
& \multicolumn{4}{c|}{\textbf{Model $S^{(0)}$ / Data $S^{(0)}$}}
& \multicolumn{4}{c|}{\textbf{Model $S^{(1)}$ / Data $S^{(01)}$}}
& \multicolumn{4}{c|}{\textbf{Model $S^{(2)}$ / Data $S^{(012)}$}}\\
\textbf{Cumulative Data} & 60.42 & 48.55 & 86.91 & 68.50
& \textbf{40.10} & 56.30 & 70.98 & \textbf{52.56}
& \textbf{21.06}& 47.56 & 55.39 & \textbf{37.33}
& \\
\bottomrule
\end{tabular}
}
\end{table*}

\begin{table*}[t]
\centering
\caption{Catastrophic forgetting comparison on MIT-States. Each model is evaluated on its session-specific test set. CCZSL-Baseline shows severe forgetting, while PromptCCZSL better preserves prior knowledge, achieving higher AUC and Attr/Obj accuracies with lower forgetting ($\mathcal{F}_{AUC}$, Eq.~\ref{eq:avg_fau}). Under cumulative zero-shot evaluation up to $S^{(t)}$, PromptCCZSL consistently outperforms the baseline.}
\label{tab:Table_7}
\resizebox{\textwidth}{!}{
\begin{tabular}{c|cccc|cccc|cccc|cccc|c}
\toprule
\multicolumn{17}{c}{\textbf{CCZSL -- Baseline}} \\
\midrule
\textbf{Data\textbackslash\ Model} 
& \multicolumn{4}{c|}{\textbf{Model $\mathcal{S}^{(0)}$}}
& \multicolumn{4}{c|}{\textbf{Model $\mathcal{S}^{(1)}$}}
& \multicolumn{4}{c|}{\textbf{Model $\mathcal{S}^{(2)}$}} 
& \multicolumn{4}{c|}{\textbf{Model $\mathcal{S}^{(3)}$}}
& $\mathcal{F}_{AUC}^{i}\downarrow$\\
& AUC & Attr & Obj & HM
& AUC & Attr & Obj & HM
& AUC & Attr & Obj & HM
& AUC & Attr & Obj & HM
&\\
\midrule
\textbf{Data $\mathcal{S}^{(0)}$} 
& 44.41 & 54.13 & 68.99 & 57.96
& 7.53 & 25.60 & 47.82 & 21.72
& 0.26 & 20.20 & 25.68 & 3.48
& 0.02 & 15.32 & 20.57 & 1.00
& \textcolor{red}{41.81}\\
\textbf{Data $\mathcal{S}^{(1)}$}
& -- & -- & -- & --
& 38.80 & 45.88 & 64.19 & 52.26
& 2.69 & 22.09 & 38.52 & 11.48
& 0.14 & 17.47 & 34.63 & 0.26
& \textcolor{red}{37.39}\\
\textbf{Data $\mathcal{S}^{(2)}$}
& -- & -- & -- & --
& -- & -- & -- & --
& 35.38 & 47.82 & 63.66 & 52.12
& 0.54 & 23.57 & 42.41 & 18.49
& \textcolor{red}{34.84}\\
\textbf{Data $\mathcal{S}^{(3)}$}
& -- & -- & -- & --
& -- & -- & -- & --
& -- & -- & -- & --
& 27.24 & 43.58 & 56.17 & 44.18
& --\\
\midrule
\multicolumn{17}{r}{}& $\mathcal{F}_{AUC}=$\textcolor{red}{38.01}\\
\midrule
& \multicolumn{4}{c|}{\textbf{Model $S^{(0)}$ / Data $S^{(0)}$}}
& \multicolumn{4}{c|}{\textbf{Model $S^{(1)}$ / Data $S^{(01)}$}}
& \multicolumn{4}{c|}{\textbf{Model $S^{(2)}$ / Data $S^{(012)}$}}
& \multicolumn{4}{c|}{\textbf{Model $S^{(3)}$ / Data $S^{(0123)}$}}\\
\textbf{Cumulative Data}
& 44.41 & 54.13 & 68.99 & 57.96
& 26.49 & 39.47 & 59.01 & 42.72
& 12.62 & 34.1 & 48.44 & 29.16
& 7.74 & 28.94 & 43.66 & 22.79  
& \\
\toprule
\multicolumn{17}{c}{\textbf{PromptCCZSL (ours)}} \\
\midrule
\textbf{Data\textbackslash\ Model} 
& \multicolumn{4}{c|}{\textbf{Model $\mathcal{S}^{(0)}$}}
& \multicolumn{4}{c|}{\textbf{Model $\mathcal{S}^{(1)}$}}
& \multicolumn{4}{c|}{\textbf{Model $\mathcal{S}^{(2)}$}} 
& \multicolumn{4}{c|}{\textbf{Model $\mathcal{S}^{(3)}$}}
& $\mathcal{F}_{AUC}^{i}\downarrow$\\
& AUC & Attr & Obj & HM
& AUC & Attr & Obj & HM
& AUC & Attr & Obj & HM
& AUC & Attr & Obj & HM
&\\
\midrule
\textbf{Data $\mathcal{S}^{(0)}$}
& 44.41 & 54.13 & 68.99 & 57.96
& 21.44 & 37.84 & 58.78 & 37.65
& 13.14 & 37.46 & 46.32 & 27.34
& 9.50 & 33.63 & 43.47 & 23.87
& \textcolor{red}{\textbf{29.72}}\\
\textbf{Data $\mathcal{S}^{(1)}$}
& -- & -- & -- & --
& 38.15 & 49.39 & 65.58 & 52.60
& 22.22 & 41.30 & 57.69 & 38.54
& 18.39 & 37.69 & 55.37 & 33.87
& \textcolor{red}{\textbf{17.85}}\\
\textbf{Data $\mathcal{S}^{(2)}$}
& -- & -- & -- & --
& -- & -- & -- & --
& 30.12 & 44.97 & 61.59 & 47.96
& 18.31 & 37.53 & 54.87 & 36.12
& \textcolor{red}{\textbf{11.81}}\\
\textbf{Data $\mathcal{S}^{(3)}$}
& -- & -- & -- & --
& -- & -- & -- & --
& -- & -- & -- & --
& 21.70 & 39.98 & 52.94 & 38.64
& --\\
\midrule
\multicolumn{17}{r}{}& $\mathcal{F}_{AUC}=$\textcolor{red}{\textbf{19.79} ($-18.22$)}\\
\midrule
& \multicolumn{4}{c|}{\textbf{Model $S^{(0)}$ / Data $S^{(0)}$}}
& \multicolumn{4}{c|}{\textbf{Model $S^{(1)}$ / Data $S^{(01)}$}}
& \multicolumn{4}{c|}{\textbf{Model $S^{(2)}$ / Data $S^{(012)}$}}
& \multicolumn{4}{c|}{\textbf{Model $S^{(3)}$ / Data $S^{(0123)}$}}\\
\textbf{Cumulative Data}
& 44.41 & 54.13 & 68.99 & 57.96
&\textbf{ 32.38} & 45.74 & 63.43 & \textbf{47.01}
& \textbf{24.05} & 42.43 & 57.69 & \textbf{41.12}
& \textbf{18.50} & 38.09 & 53.08 & \textbf{35.23 } 
& \\
\bottomrule
\end{tabular}
}
\end{table*}
\begin{table*}[t]
\centering
\caption{Catastrophic forgetting comparison on MIT-States. Each model is evaluated on its session-specific test set. CCZSL-Baseline shows severe forgetting, while PromptCCZSL better preserves prior knowledge, achieving higher AUC and Attr/Obj accuracies with lower forgetting ($\mathcal{F}_{AUC}$, Eq.~\ref{eq:avg_fau}). Under cumulative zero-shot evaluation up to $S^{(t)}$, PromptCCZSL consistently outperforms the baseline.} 
\label{tab:Table_8}
\resizebox{\textwidth}{!}{
\begin{tabular}{c|cccc|cccc|cccc|cccc|cccc|c}
\toprule
\multicolumn{22}{c}{\textbf{CCZSL -- Baseline}} \\
\midrule
\textbf{Data\textbackslash\ Model} 
& \multicolumn{4}{c|}{\textbf{Model $\mathcal{S}^{(0)}$}}
& \multicolumn{4}{c|}{\textbf{Model $\mathcal{S}^{(1)}$}}
& \multicolumn{4}{c|}{\textbf{Model $\mathcal{S}^{(2)}$}} 
& \multicolumn{4}{c|}{\textbf{Model $\mathcal{S}^{(3)}$}}
& \multicolumn{4}{c|}{\textbf{Model $\mathcal{S}^{(4)}$}}
& $\mathcal{F}_{AUC}^{i}\downarrow$\\
& AUC & Attr & Obj & HM
& AUC & Attr & Obj & HM
& AUC & Attr & Obj & HM
& AUC & Attr & Obj & HM
& AUC & Attr & Obj & HM
&\\
\midrule
\textbf{Data $\mathcal{S}^{(0)}$} 
& 16.58 & 45.59 & 62.32 & 34.57
& 0.89 & 17.19 & 20.91 & 6.69
& 0.04 & 0.43 & 29.47 & 0.12
& 0.01 & 4.18 & 29.88 & 0.48
& 0.02 & 5.51 & 25.09 & 0.61
& \textcolor{red}{16.34}\\
\textbf{Data $\mathcal{S}^{(1)}$}
& -- & -- & -- & --
& 16.51 & 37.87 & 66.27 & 34.41
& 3.76 & 17.46 & 45.56 & 15.38
& 1.63 & 13.31 & 39.64 & 0.94
& 0.09 & 9.76 & 29.88 & 7.10
& \textcolor{red}{14.68}\\
\textbf{Data $\mathcal{S}^{(2)}$}
& -- & -- & -- & --
& -- & -- & -- & --
& 16.64 & 68.35 & 63.79 & 34.31
& 6.32 & 41.69 & 52.14 & 20.68
& 3.99 & 36.02 & 46.89 & 16.44
& \textcolor{red}{16.65}\\
\textbf{Data $\mathcal{S}^{(3)}$}
& -- & -- & -- & --
& -- & -- & -- & --
& -- & -- & -- & --
& 11.57 & 52.62 & 53.92 & 29.87
& 7.15 & 41.33 & 37.03 & 22.24
& \textcolor{red}{4.42}\\
\textbf{Data $\mathcal{S}^{(4)}$}
& -- & -- & -- & --
& -- & -- & -- & --
& -- & -- & -- & --
& -- & -- & -- & --
& 19.96 & 52.02 & 79.57 & 37.87
&--\\
\midrule
\multicolumn{21}{r}{} & $\mathcal{F}_{AUC}=$\textcolor{red}{\textbf{13.02}}\\
\midrule
& \multicolumn{4}{c|}{\textbf{Model $S^{(0)}$ / Data $S^{(0)}$}}
& \multicolumn{4}{c|}{\textbf{Model $S^{(1)}$ / Data $S^{(01)}$}}
& \multicolumn{4}{c|}{\textbf{Model $S^{(2)}$ / Data $S^{(012)}$}}
& \multicolumn{4}{c|}{\textbf{Model $S^{(3)}$ / Data $S^{(0123)}$}}
& \multicolumn{4}{c|}{\textbf{Model $S^{(4)}$ / Data $S^{(01234)}$}}\\
\textbf{Cumulative Data}
& 16.58 & 45.59 & 62.32 & 34.57
& 2.16 &  20.23 & 27.58 & 11.19
& 3.82 & 25.47 & 41.17 & 16.51
& 2.68 & 24.84 & 41.19 & 13.41
& 2.75 & 24.0 & 37.5 & 13.81
&\\
\toprule
\multicolumn{22}{c}{\textbf{PromptCCZSL (ours)}} \\
\midrule
\textbf{Data\textbackslash\ Model} 
& \multicolumn{4}{c|}{\textbf{Model $\mathcal{S}^{(0)}$}}
& \multicolumn{4}{c|}{\textbf{Model $\mathcal{S}^{(1)}$}}
& \multicolumn{4}{c|}{\textbf{Model $\mathcal{S}^{(2)}$}} 
& \multicolumn{4}{c|}{\textbf{Model $\mathcal{S}^{(3)}$}}
& \multicolumn{4}{c|}{\textbf{Model $\mathcal{S}^{(4)}$}}
& $\mathcal{F}_{AUC}^{i}\downarrow$\\
& AUC & Attr & Obj & HM
& AUC & Attr & Obj & HM
& AUC & Attr & Obj & HM
& AUC & Attr & Obj & HM
& AUC & Attr & Obj & HM
&\\
\hline
\textbf{Data $\mathcal{S}^{(0)}$}
& 16.58 & 45.59 & 62.32 & 34.57
& 13.97 & 43.19 & 57.27 & 31.32
& 11.94 & 37.63 & 56.04 & 28.54
& 11.03 & 37.79 & 54.51 & 27.24
& 9.22 & 34.39 & 53.14 & 25.29
& \textcolor{red}{\textbf{5.04}}\\
\textbf{Data $\mathcal{S}^{(1)}$}
& -- & -- & -- & --
& 16.86 & 40.83 & 61.24 & 34.43
& 12.78 & 34.91 & 56.80 & 29.44
& 10.47 & 35.21 & 54.73 & 25.81
& 9.11 & 30.18 & 48.52 & 24.34
& \textcolor{red}{\textbf{6.07}}\\
\textbf{Data $\mathcal{S}^{(2)}$}
& -- & -- & -- & --
& -- & -- & -- & --
& 11.99 & 58.74 & 63.50 & 27.19
& 10.63 & 47.77 & 58.93 & 26.39
& 10.44 & 45.24 & 56.80 & 26.20
& \textcolor{red}{\textbf{2.14}}\\
\textbf{Data $\mathcal{S}^{(3)}$}
& -- & -- & -- & --
& -- & -- & -- & --
& -- & -- & -- & --
& 12.01 & 44.82 & 49.33 & 29.43
& 7.35 & 31.90 & 44.41 & 22.39
& \textcolor{red}{\textbf{4.66}}\\
\textbf{Data $\mathcal{S}^{(4)}$}
& -- & -- & -- & --
& -- & -- & -- & --
& -- & -- & -- & --
& -- & -- & -- & --
& 13.48 & 36.82 & 69.60 & 30.03
& --\\
\midrule
\multicolumn{21}{r}{} & $\mathcal{F}_{AUC}=$\textcolor{red}{\textbf{4.48} ($-8.54$)}\\
\midrule
& \multicolumn{4}{c|}{\textbf{Model $S^{(0)}$ / Data $S^{(0)}$}}
& \multicolumn{4}{c|}{\textbf{Model $S^{(1)}$ / Data $S^{(01)}$}}
& \multicolumn{4}{c|}{\textbf{Model $S^{(2)}$ / Data $S^{(012)}$}}
& \multicolumn{4}{c|}{\textbf{Model $S^{(3)}$ / Data $S^{(0123)}$}}
& \multicolumn{4}{c|}{\textbf{Model $S^{(4)}$ / Data $S^{(01234)}$}}\\
\textbf{Cumulative Data}
& 16.58 & 45.59 & 62.32 & 34.57
& \textbf{14.53} & 41.54 & 57.24 & \textbf{32.02}
& \textbf{13.66} &  43.89 & 58.43 & \textbf{30.94}
& \textbf{11.44} & 39.94 & 53.86 & \textbf{28.09}
& \textbf{9.83} & 35.01 & 52.28 & \textbf{25.98}
&\\
\bottomrule
\end{tabular}}
\end{table*}
\subsection{Ablations Studies}
\label{subsec:ablation}
\subsubsection{Ablation on Major Components} 

Tab.~\ref{tab:Table_9} presents the component-wise ablation study. The baseline using only $\mathcal{L}_{\text{CSKD}}$ without SAwM$^2$F achieves $15.39\%$ and $1.61\%$ AUC at $S^{(1)}$ and $S^{(2)}$, respectively. Adding SAwM$^2$F improves AUC to $34.50\%$ and $17.74\%$, showing its strong effect in reducing catastrophic forgetting.
Adding $\mathcal{L}_{\text{PAAL}}$ further increases AUC to $36.90\%$ and $20.72\%$, while $\mathcal{L}_{\text{OPL}}$ improves object preservation, reaching $70.02\%$ object accuracy at $S^{(1)}$. The full model with $\mathcal{L}_{\text{IDL}}$ achieves the best overall performance, with $40.10\%$ and $21.06\%$ AUC at $S^{(1)}$ and $S^{(2)}$, respectively.
Overall, SAwM$^2$F provides the largest gain, while PAAL, OPL, and IDL offer complementary improvements in alignment, semantic preservation, and representation diversity.

\subsubsection{Ablation on Hyperparameters}

We evaluate the sensitivity of the proposed framework to key hyperparameters on the validation set. Unless otherwise stated, hyperparameters are selected using a single validation session and then fixed across all continual sessions, demonstrating good transferability without session-specific tuning. Overall, the ablation results demonstrate that the proposed framework is not highly sensitive to hyperparameter selection and achieves stable performance across a broad range of settings.
Effect of CSKD and compositional loss weights.
The coefficients $\lambda_{kd}$ and $\lambda_{ce}$ control the balance between preserving previously acquired knowledge and adapting to newly introduced compositions. As shown in Tab.~\ref{tab:Table_10}, assigning a larger weight to the distillation objective generally improves compositional retention. Among the evaluated settings, $\lambda_{kd}=0.65$ and $\lambda_{ce}=0.35$ achieve the highest validation AUC (23.56\%), indicating the most favorable stability--plasticity trade-off. The comparable performance obtained with linearly decayed weights further suggests that the framework is relatively robust to moderate variations of these coefficients.
Effect of the Orthogonal Projection Loss (OPL).
Tab.~\ref{tab:Table_11} investigates the influence of the OPL weight $\lambda_{opl}$. Increasing $\lambda_{opl}$ from 0.01 to 0.05 substantially improves all evaluation metrics, boosting AUC ($43.74\% \rightarrow 51.22\%$) and compositional accuracy ($36.40\%\rightarrow41.40\%$. Similar gains are observed for attribute and object recognition accuracy. These results indicate that an appropriate OPL weight effectively preserves object semantics and reduces cross-session semantic interference, leading to a more discriminative compositional embedding space. In contrast, decaying schedule results in inferior performance, suggesting that a consistently enforced object-preserving constraint is beneficial throughout training.
Effect of the Intra-Session Diversification Loss (IDL).
Tab.~\ref{tab:Table_12} evaluates different values of $\lambda_{opl}$ while keeping $\lambda_{idl}=0.05$ yielding the highest AUC ($46.26\%$) and compositional accuracy ($36.37\%$).

\subsubsection{SAwM2F Updating Strategies} Tab.~\ref{tab:Table_13} shows that updating only the current session's tail embeddings improves AUC performance ($+23.46\%$), while updating both head (prior session embeddings) and tail degrades it, emphasizing the need to preserve prior-session embeddings.

\subsubsection{Orthogonality Regularizers Choice}
Tab.~\ref{tab:Table_14} compares different regularization strategies in the absence of session awareness. Using single regularizer underperforms the full configuration. Combining OPL (Eq.~\ref{eq: opl}) and IDL (Eq. \ref{eq: idl}) yields the best results, highlighting the need to jointly enforce retention and adaptive plasticity for robust continual CCZSL.
\begin{table}[t]
\centering
\caption{Component-wise ablation of PromptCCZSL on UT-Zappos. Starting from the baseline (CSKD--SAwM2F), PAAL, OPL, and IDL are added incrementally. Results for $S^{(1)}$ and $S^{(2)}$ report AUC, Attr, and Obj accuracies under cumulative evaluation, with the full model achieving the best performance (Sec.~\ref{subsec:ablation}).}
\label{tab:Table_9}
\resizebox{0.6\textwidth}{!}{
\setlength{\tabcolsep}{4.5pt}
\renewcommand{\arraystretch}{0.95}
\begin{tabular}{l|ccc|ccc}
\toprule
\textbf{Components} & \multicolumn{3}{c|}{\textbf{$S^{(1)}$ Model / $S^{(01)}$ Data}} & \multicolumn{3}{c}{\textbf{$S^{(2)}$ Model / $S^{(012)}$ Data}}\\
\cmidrule(lr){2-4}\cmidrule(lr){5-7}
& \textbf{AUC} $\uparrow$ & \textbf{Attr} $\uparrow$ & \textbf{Obj} $\uparrow$ & \textbf{AUC} $\uparrow$ & \textbf{Attr} $\uparrow$ & \textbf{Obj} $\uparrow$\\
\midrule
$\mathcal{L}_{\mathrm{CSKD}} - \mathrm{SAwM2F}$  & 15.39 & 49.02 & 43.28 &  1.61 & 37.47 & 30.30\\
$\mathcal{L}_{\mathrm{CSKD}} + \mathrm{SAwM2F}$             & 34.50 & 54.22 & 65.41 & 17.74 & 45.33 & 52.09\\
\ $+\mathcal{L}_{\mathrm{PAAL}}$  & 36.90 & 53.85 & 69.90 & 20.72 & 45.54 & \textbf{56.42}\\
\ $+\mathcal{L}_{\mathrm{OPL}}$  & 39.34 & 54.22 & 70.02 & 17.04 & 43.82 & 53.29\\
\ $+\mathcal{L}_{\mathrm{IDL}}$  & \textbf{40.10} & \textbf{56.30} & \textbf{70.98} & \textbf{21.06} & \textbf{47.56} & 55.39\\
\bottomrule
\end{tabular}
}
\end{table}
\begin{table}[t]
\centering
\caption{Ablation of CSKD ($\lambda_{kd}$) and compositional loss ($\lambda_{ce}$) weights on the UT-Zappos~\cite{yu2014fine} validation set, showing the retention–adaptation trade-off.}
\label{tab:Table_10}
\resizebox{0.55\columnwidth}{!}{
\begin{tabular}{cc|cccc}
\toprule
\textbf{$\lambda_{kd}$} & \textbf{$\lambda_{ce}$} &
\multicolumn{4}{c}{\textbf{\textbf{$S^{(1)}$ Model / $S^{(01)}$ Data}}}\\
 & & AUC $\uparrow$ & Comp $\uparrow$ & Attr $\uparrow$ & Obj $\uparrow$\\
\hline
0.50 & 1.00 & 23.17 & 23.38 & 39.40 & 61.67\\
0.65 & 1.00 & 22.60 & 22.06 & 39.17 & 61.11\\
\textbf{0.65} & \textbf{0.35} & 23.56 & 23.17 & 38.05 & 59.82\\
0.65--0.35 & 0.35--0.65 & 23.50 & 23.17 & 38.05 & 59.82\\
\hline
\end{tabular}}
\end{table}

\begin{table}[t]
\centering
\caption{Ablation of Orthogonal Projection Loss weight ($\lambda_{\mathrm{opl}}$) on UT-Zappos~\cite{yu2014fine}, with $\lambda_{\mathrm{opl}}=0.05$ achieving the best retention–adaptation balance.
}
\label{tab:Table_11}
\resizebox{0.45\columnwidth}{!}{
\begin{tabular}{c|cccc}
\hline
\textbf{$\lambda_{opl}$} &
\multicolumn{4}{c}{\textbf{$S^{(1)}$ Model / $S^{(01)}$ Data}}\\
\hline
 & AUC $\uparrow$ & Comp $\uparrow$ & Attr $\uparrow$ & Obj $\uparrow$\\
\hline
0.01 & 43.74 & 36.4 & 49.86 & 74.76 \\
\textbf{0.05} & \textbf{51.22} & \textbf{41.4} & \textbf{53.11} & \textbf{77.33}  \\
decay$_{[0.01]\downarrow}$ & 44.07 & 37.66 & 50.03 & 75.32 \\
\hline
\end{tabular}}
\end{table}
\begin{table}[t]
\centering
\caption{Ablation of Intra-Session Diversification Loss weight ($\lambda_{\mathrm{idl}}$=0.005) on UT-Zappos, showing the trade-off between compositional diversity and representation stability.}
\label{tab:Table_12}
\resizebox{0.45\columnwidth}{!}{
\begin{tabular}{c|cccc}
\hline
\textbf{$\lambda_{opl}$} &
\multicolumn{4}{c}{\textbf{$S^{(1)}$ Model / $S^{(01)}$ Data}}\\
\hline
 & AUC $\uparrow$ & Comp $\uparrow$ & Attr $\uparrow$ & Obj $\uparrow$\\
\hline
0.01 & 39.44 & 35.25 & 48.18 & 74.43 \\
0.05 & \textbf{46.26} & \textbf{36.37} & \textbf{48.63} & \textbf{75.94} \\
decay$_{[0.01]\downarrow}$ & 35.7 & 34.30 & 46.28 & 76.83 \\
\hline
\end{tabular}}
\end{table}
\begin{table*}[t]
\centering
\caption{Ablation of head and tail composition updates in SAwM2F on UT-Zappos, showing that freezing the head and updating the tail yields the best continual generalization.}
\label{tab:Table_13}
\resizebox{0.45\columnwidth}{!}{
\begin{tabular}{cc|cccc}
\hline
head & tail &
\multicolumn{4}{c}{\textbf{$S^{(1)}$ Model / $S^{(01)}$ Data}}\\
\hline
 & & AUC $\uparrow$ & Comp $\uparrow$ & Attr $\uparrow$ & Obj $\uparrow$\\
\hline
\cmark & \cmark  & 23.69 & 22.89 & 37.55 & 60.04 \\
\xmark & \cmark & \textbf{47.15} & \textbf{38.33} & \textbf{49.58} & \textbf{77.11} \\
\hline
\end{tabular}}
\end{table*}
\begin{table}[t]
    \centering
    \caption{Comparison of orthogonality regularization on UT-Zappos. Combining Orthogonal Projection Loss (OPL) and Intra-Session Diversification Loss (IDL) improves compositional performance by promoting inter-session separation and intra-session diversity.}
    \label{tab:Table_14}
    \resizebox{0.45\columnwidth}{!}{
    \begin{tabular}{lccc}
    \toprule
    \textbf{$S^{(1)}$ Model / Data} & \textbf{$S^{(0)}$} & \textbf{$S^{(1)}$} & \textbf{$S^{(01)}$} \\
    \midrule
    w/ Single Regularizer   & 34.64 & \textbf{36.82} & 38.42 \\
    w/ OPL+IDL (PromptCCZSL)       & \textbf{37.35} & 34.82 & \textbf{40.1} \\
    \hline
    \end{tabular}}
\end{table}
\begin{table}[t]
\centering
\caption{
Model size evolution across sessions.
$P(t)=P_{\text{const}}+P_{\text{B}}(t)$.
Only the soft prompt bank grows, while all other modules maintain constant parameters complexity.
}
\label{tab:Table_15}
\resizebox{0.7\textwidth}{!}{
\begin{tabular}{lcc}
\hline
\textbf{Component} &
\textbf{Status} &
\textbf{Parameter Growth ($P$)} \\
\hline

Vision Encoder (w/ adapters)
& Fixed (Learnable)
& Constant \\

Text Encoder
& Fixed (Frozen)
& Constant \\

Fusion Module (SAwM2F/SAgM2F)
& Fixed (Learnable)
& Constant \\

Disentanglers
& Fixed (Learnable)
& Constant \\

Soft Prompt Bank
& Expanding
& $B(t-1)+\Delta B(t)$ \\
\hline

\end{tabular}
}
\end{table}

\begin{table}[t]
\centering
\caption{PromptCCZSL efficiency comparison within $S^{(t)}$ sessions. Total and trainable parameters are reported (in millions), together with GPU memory usage.}
\label{tab:Table_16}
\resizebox{0.7\textwidth}{!}{%
\begin{tabular}{ll|ccc|ccccc}
\toprule
\multirow{2}{*}{Model} &
\multirow{2}{*}{\#Params} &
\multicolumn{3}{c|}{\textbf{UT-Zappos}} &
\multicolumn{5}{c}{\textbf{C-GQA}} \\
\cmidrule(lr){3-5}
\cmidrule(l){6-10}
&&
$S^{(0)}$ & $S^{(1)}$ & $S^{(2)}$
&
$S^{(0)}$ & $S^{(1)}$ & $S^{(2)}$ & $S^{(3)}$ & $S^{(4)}$\\
\midrule

$\mathbf{S^{(t-1)}}$
& Total
& 449.3 & 449.3 & 449.3
& 449.7 & 450.0 & 450.0 & 450.0 & 450.0\\

& Trainable
& 21.7 & 0 & 0
& 22.2 & 0 & 0 & 0 & 0\\

$\mathbf{S^{(t)}}$
& Total
& -- & 449.3 & 449.3
& -- & 449.7 & 449.8 & 449.9 & 450.0\\

& Trainable
& -- & 21.7 & 21.7
& -- & 22.2 & 22.3 & 22.4 & 22.5\\

\midrule

\multicolumn{2}{l|}{GPU Memory (GB)}
& 12 & 14 & 16
& 18 & 24 & 28 & 32 & 32\\

\bottomrule
\end{tabular}%
}
\end{table}
\subsubsection{Scalability Analysis}
\begin{figure}
    \centering
    \includegraphics[width=1.0\linewidth]{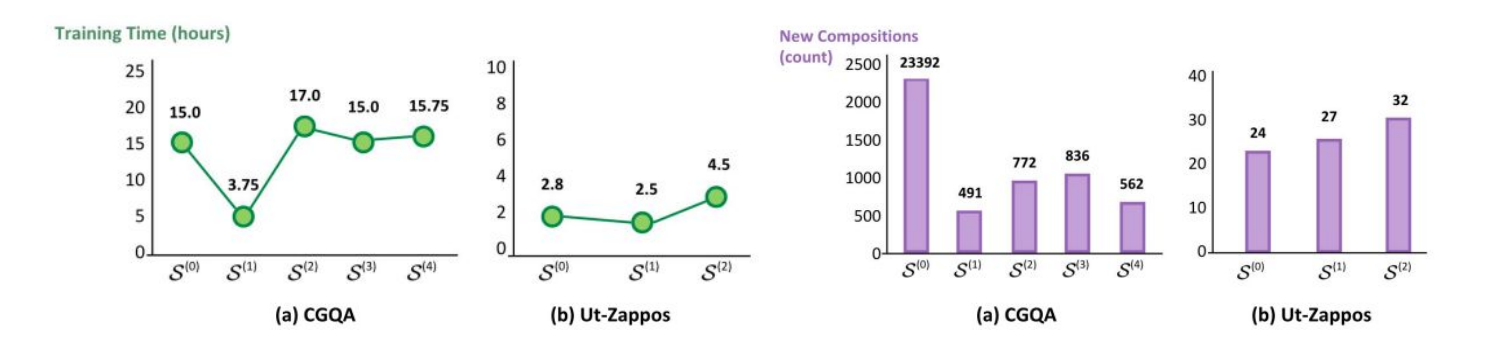}
    \caption{\textbf{Training time} across C-GQA and UT-Zappos sessions on a single NVIDIA V100 32GB GPU. Despite varying numbers of new compositions, training time remains relatively stable, demonstrating PromptCCZSL's scalability.}
    \label{fig:rebut_training_time}
\end{figure}

We evaluate scalability from three perspectives: parameter growth (Tab.~\ref{tab:Table_15}), parameter and memory overhead (Tab.~\ref{tab:Table_16}), and training time (Fig.~\ref{fig:rebut_training_time}). 

\subsection{Qualitative Results}

\label{abl_sec}
\subsubsection{t-SNE Visualization}
\label{abl_subsec}

\begin{figure*}
    \centering
    \includegraphics[width=1.0\linewidth]{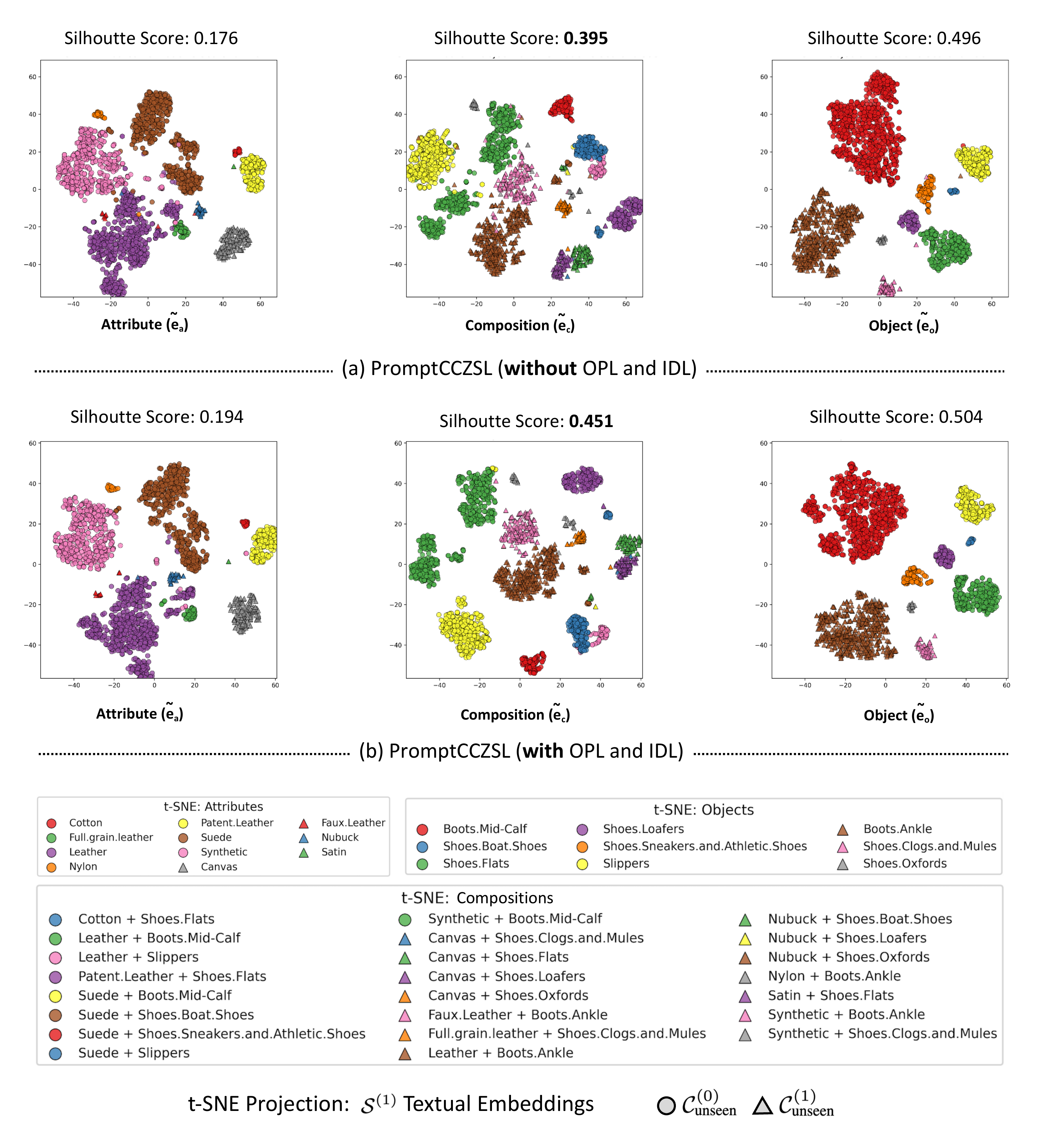}
    \caption{
    \textbf{t-SNE Projections of Textual Embeddings} from the Session-1 model on the UT-Zappos~\cite{yu2014fine} Session-01 Unseen dataset for PromptCCZSL-Troika. (a) The variant integrated Cross-Session Knowledge Distillation (CSKD), Primitive-aware Anchor Alignment Loss (PAAL), and Session-Aware Multi-Modal Fusion (SAwM2F) modules exhibits overlapping attribute, object, and composition clusters, yielding lower silhouette scores \cite{shahapure2020cluster}. (b) Incorporating Orthogonal Projection Loss (OPL) and Intra-session Diversification Loss (IDL) produces more clearly separated clusters, increases silhouette scores, and mitigates catastrophic forgetting by enforcing orthogonality between new-session embeddings and previously learned representations. Average silhouette scores summarizes overall clustering quality (ranging from --1 to 1, with higher values indicating better cluster separation).
    }
    \label{fig:abl_tsne}
\end{figure*}

To analyze module effects on the representation space, we use t-SNE and silhouette scores~\cite{shahapure2020cluster}, where higher scores indicate better separation. Fig.~\ref{fig:abl_tsne} shows Session-1 textual embeddings. In (a), CSKD, PAAL, and SAwM2F retain previous-session clusters but exhibit overlap among embeddings. Adding OPL and IDL in (b) yields cleaner clusters and higher silhouette scores, highlighting their role in preventing embedding collapse and preserving prior-session knowledge.

\subsection{Qualitative Results}
\begin{figure*}
    \centering
    \includegraphics[width=0.9\linewidth]{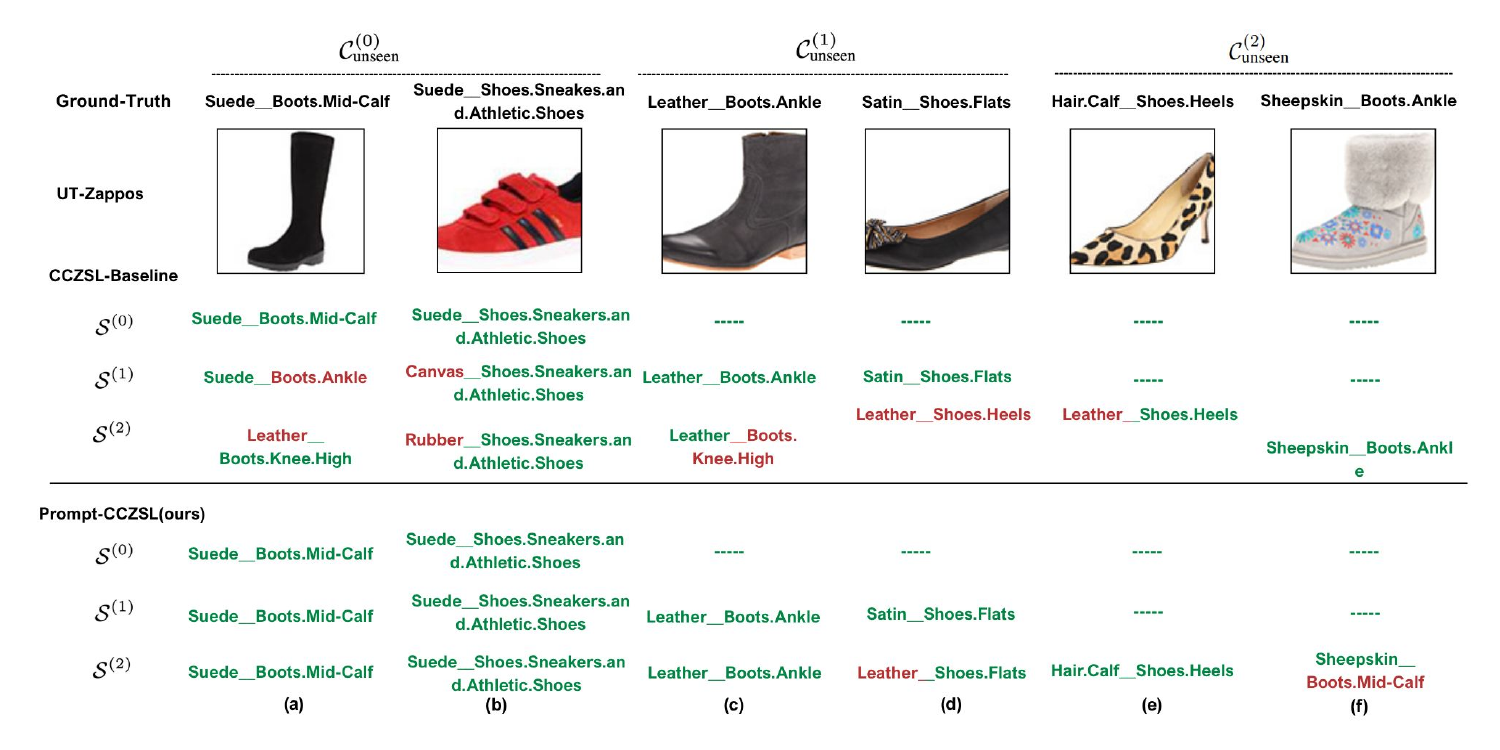}
    \caption{Qualitative comparison on UT-Zappos. Prompt-CCZSL (bottom) consistently preserves previously learned unseen compositions while adapting to newly introduced ones, outperforming the CCZSL-Baseline (top), which exhibits progressive catastrophic forgetting across sessions. Session models $S^{(0)}$, $S^{(1)}$, and $S^{(2)}$ are evaluated on unseen composition test sets $C_{\text{unseen}}^{(0)}$ (a--b), $C_{\text{unseen}}^{(1)}$ (c--d), and $C_{\text{unseen}}^{(2)}$ (e--f). Specifically, $C_{\text{unseen}}^{(0)}$ is evaluated using $S^{(0)}$, $S^{(1)}$, and $S^{(2)}$; $C_{\text{unseen}}^{(1)}$ using $S^{(1)}$ and $S^{(2)}$; and $C_{\text{unseen}}^{(2)}$ using $S^{(2)}$ only. Correct predictions are shown in \textcolor{green}{green}, while incorrect predictions are shown in \textcolor{red}{red}.}
    \label{fig:utzappos_qres}
\end{figure*}

\begin{figure*}
    \centering
    \includegraphics[width=0.9\linewidth]{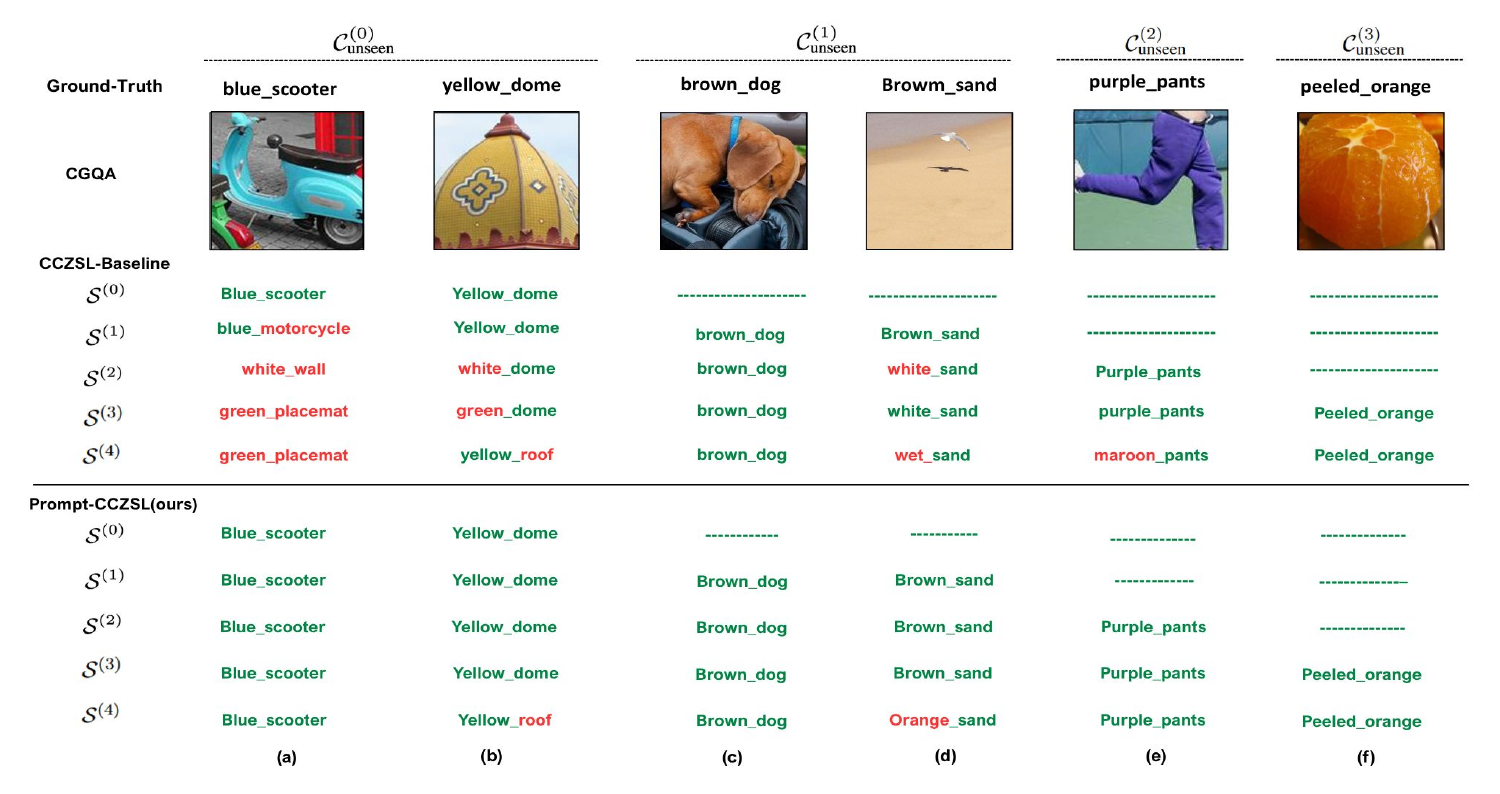}
   \caption{Qualitative comparison on CGQA. Prompt-CCZSL (bottom) consistently preserves previously learned unseen compositions while adapting to newly introduced ones, outperforming the CCZSL-Baseline (top), which exhibits progressive catastrophic forgetting across sessions. Session models $S^{(0)}$, $S^{(1)}$, $S^{(2)}$, $S^{(3)}$, and $S^{(4)}$ are evaluated on unseen composition test sets $C_{\text{unseen}}^{(0)}$ (a--b), $C_{\text{unseen}}^{(1)}$ (c--d), $C_{\text{unseen}}^{(2)}$ (e), and $C_{\text{unseen}}^{(3)}$ (f). Specifically, $C_{\text{unseen}}^{(0)}$ is evaluated using $S^{(0)}$, $S^{(1)}$, $S^{(2)}$, $S^{(3)}$, and $S^{(4)}$; $C_{\text{unseen}}^{(1)}$ using $S^{(1)}$, $S^{(2)}$, $S^{(3)}$, and $S^{(4)}$; $C_{\text{unseen}}^{(2)}$ using $S^{(2)}$, $S^{(3)}$, and $S^{(4)}$; and $C_{\text{unseen}}^{(3)}$ using $S^{(3)}$ and $S^{(4)}$. Correct predictions are shown in \textcolor{green}{green}, while incorrect predictions are shown in \textcolor{red}{red}.}
    \label{fig:cgqa_qres}
\end{figure*}

Figures~\ref{fig:utzappos_qres}, and \ref{fig:cgqa_qres} compare \textbf{CCZSL-Baseline} and \textbf{Prompt-CCZSL} across UT-Zappos, and CGQA. Each figure evaluates session models $S^{(0)}, S^{(1)}, \ldots$ on cumulative unseen composition sets $C_{\text{unseen}}^{(t)}$.
Across all datasets, \textbf{CCZSL-Baseline} shows pronounced catastrophic forgetting. While it often recognizes compositions introduced in the current session, its predictions for previously learned compositions progressively deteriorate, frequently being replaced by semantically related but incorrect attribute--object pairs. In contrast, \textbf{Prompt-CCZSL} preserves earlier compositional knowledge while adapting to newly introduced compositions.
On UT-Zappos (Fig.~\ref{fig:utzappos_qres}), Prompt-CCZSL retains correct predictions for previously learned shoe-attribute compositions across sessions. Similar behavior is observed on CGQA (Fig.~\ref{fig:cgqa_qres}), where it consistently recognizes compositions such as \textit{blue\_scooter}, \textit{yellow\_dome}, and \textit{brown\_dog}. For \textit{brown\_sand}, the baseline progressively drifts from \textit{brown\_sand} to \textit{white\_sand} and \textit{wet\_sand}, whereas Prompt-CCZSL retains the correct composition across sessions and produces the semantically closer \textit{orange\_sand} in the final session.

\section{Limitations}
\label{abl_sec:limitations}
 
Our experiments were conducted using an NVIDIA V100 with 32 GB memory. As CCZSL sessions scale, memory and compute demands rise, making later sessions costly on mid-range GPUs and is a potential focus for future work.

\section{Conclusion}
In this work, we introduce PromptCCZSL, the first prompt-based framework for continual compositional zero-shot learning in vision-language models. By combining session-aware compositional prompting, semantic anchoring, cross-session distillation, and orthogonality regularization, it mitigates catastrophic forgetting while preserving primitive-level consistency as new compositions emerge. Unlike static CZSL methods, PromptCCZSL enables scalable incremental adaptation. Extensive evaluations on UT-Zappos, C-GQA and MIT-States demonstrate consistent and substantial improvements over strong VLM and non-VLM baselines.

\bibliographystyle{elsarticle-num}
\bibliography{refs}

@article{huynh2020compositional,
  title={Compositional zero-shot learning via fine-grained dense feature composition},
  author={Huynh, Dat and Elhamifar, Ehsan},
  journal={Advances in Neural Information Processing Systems},
  volume={33},
  pages={19849--19860},
  year={2020}
}

@inproceedings{li2022siamese,
  title={Siamese contrastive embedding network for compositional zero-shot learning},
  author={Li, Xiangyu and Yang, Xu and Wei, Kun and Deng, Cheng and Yang, Muli},
  booktitle={Proceedings of the IEEE/CVF conference on computer vision and pattern recognition},
  pages={9326--9335},
  year={2022}
}

@inproceedings{naeem2021learning,
  title={Learning graph embeddings for compositional zero-shot learning},
  author={Naeem, Muhammad Ferjad and Xian, Yongqin and Tombari, Federico and Akata, Zeynep},
  booktitle={Proceedings of the IEEE/CVF conference on computer vision and pattern recognition},
  pages={953--962},
  year={2021}
}

@inproceedings{zhang2022learning,
  title={Learning invariant visual representations for compositional zero-shot learning},
  author={Zhang, Tian and Liang, Kongming and Du, Ruoyi and Sun, Xian and Ma, Zhanyu and Guo, Jun},
  booktitle={European conference on computer vision},
  pages={339--355},
  year={2022},
  organization={Springer}
}

@inproceedings{radford2021learning,
  title={Learning transferable visual models from natural language supervision},
  author={Radford, Alec and Kim, Jong Wook and Hallacy, Chris and Ramesh, Aditya and Goh, Gabriel and Agarwal, Sandhini and Sastry, Girish and Askell, Amanda and Mishkin, Pamela and Clark, Jack and others},
  booktitle={International conference on machine learning},
  pages={8748--8763},
  year={2021},
  organization={PmLR}
}

@inproceedings{lu2023decomposed,
  title={Decomposed soft prompt guided fusion enhancing for compositional zero-shot learning},
  author={Lu, Xiaocheng and Guo, Song and Liu, Ziming and Guo, Jingcai},
  booktitle={Proceedings of the IEEE/CVF Conference on Computer Vision and Pattern Recognition},
  pages={23560--23569},
  year={2023}
}

@inproceedings{nayaklearning,
  title={Learning to Compose Soft Prompts for Compositional Zero-Shot Learning},
  author={Nayak, Nihal V and Yu, Peilin and Bach, Stephen},
  booktitle={The Eleventh International Conference on Learning Representations},
  pages={1},
  year={2023}
}

@inproceedings{huang2024troika,
  title={Troika: Multi-path cross-modal traction for compositional zero-shot learning},
  author={Huang, Siteng and Gong, Biao and Feng, Yutong and Zhang, Min and Lv, Yiliang and Wang, Donglin},
  booktitle={Proceedings of the IEEE/CVF Conference on Computer Vision and Pattern Recognition},
  pages={24005--24014},
  year={2024}
}

@article{hu2022lora,
  title={Lora: Low-rank adaptation of large language models.},
  author={Hu, Edward J and Shen, Yelong and Wallis, Phillip and Allen-Zhu, Zeyuan and Li, Yuanzhi and Wang, Shean and Wang, Liang and Chen, Weizhu and others},
  journal={Iclr},
  volume={1},
  number={2},
  pages={3},
  year={2022}
}

@article{karimi2021compacter,
  title={Compacter: Efficient low-rank hypercomplex adapter layers},
  author={Karimi Mahabadi, Rabeeh and Henderson, James and Ruder, Sebastian},
  journal={Advances in neural information processing systems},
  volume={34},
  pages={1022--1035},
  year={2021}
}

@article{rebuffi2017learning,
  title={Learning multiple visual domains with residual adapters},
  author={Rebuffi, Sylvestre-Alvise and Bilen, Hakan and Vedaldi, Andrea},
  journal={Advances in neural information processing systems},
  volume={30},
  year={2017}
}

@inproceedings{
    jung2023new,
    title={New Insights for the Stability-Plasticity Dilemma in Online Continual Learning},
    author={Dahuin Jung and Dongjin Lee and Sunwon Hong and Hyemi Jang and Ho Bae and Sungroh Yoon},
    booktitle={International Conference on Learning Representations},
    pages={1},
    year={2023},
}

@article{li2017learning,
  title={Learning without forgetting},
  author={Li, Zhizhong and Hoiem, Derek},
  journal={IEEE transactions on pattern analysis and machine intelligence},
  volume={40},
  number={12},
  pages={2935--2947},
  year={2017},
  publisher={IEEE}
}

@inproceedings{misra2017red,
  title={From red wine to red tomato: Composition with context},
  author={Misra, Ishan and Gupta, Abhinav and Hebert, Martial},
  booktitle={Proceedings of the IEEE conference on computer vision and pattern recognition},
  pages={1792--1801},
  year={2017}
}

@inproceedings{nagarajan2018attributes,
  title={Attributes as operators: factorizing unseen attribute-object compositions},
  author={Nagarajan, Tushar and Grauman, Kristen},
  booktitle={European Conference on Computer Vision},
  pages={172--190},
  year={2018},
  organization={Springer}
}

@inproceedings{li2020symmetry,
  title={Symmetry and group in attribute-object compositions},
  author={Li, Yong-Lu and Xu, Yue and Mao, Xiaohan and Lu, Cewu},
  booktitle={2020 IEEE/CVF Conference on Computer Vision and Pattern Recognition (CVPR)},
  pages={11313--11322},
  year={2020},
  organization={IEEE}
}

@article{jiang2024mutual,
  title={Mutual balancing in state-object components for compositional zero-shot learning},
  author={Jiang, Chenyi and Ye, Qiaolin and Wang, Shidong and Shen, Yuming and Zhang, Zheng and Zhang, Haofeng},
  journal={Pattern Recognition},
  volume={152},
  pages={110451},
  year={2024},
  publisher={Elsevier}
}

@inproceedings{wang2023learning,
  title={Learning conditional attributes for compositional zero-shot learning},
  author={Wang, Qingsheng and Liu, Lingqiao and Jing, Chenchen and Chen, Hao and Liang, Guoqiang and Wang, Peng and Shen, Chunhua},
  booktitle={Proceedings of the IEEE/CVF conference on computer vision and pattern recognition},
  pages={11197--11206},
  year={2023}
}

@article{shuang2025visual,
  title={Visual primitives as words: Alignment and interaction for compositional zero-shot learning},
  author={Shuang, Feng and Li, Jiahuan and Huang, Qingbao and Zhao, Wenye and Xu, Dongsheng and Han, Chao and Cheng, Haonan},
  journal={Pattern Recognition},
  volume={157},
  pages={110814},
  year={2025},
  publisher={Elsevier}
}

@inproceedings{wang2023hierarchical,
  title={Hierarchical Prompt Learning for Compositional Zero-Shot Recognition.},
  author={Wang, Henan and Yang, Muli and Wei, Kun and Deng, Cheng},
  booktitle={IJCAI},
  pages={1470--1478},
 year={2023}
}

@article{kirkpatrick2017overcoming,
  title={Overcoming catastrophic forgetting in neural networks},
  author={Kirkpatrick, James and Pascanu, Razvan and Rabinowitz, Neil and Veness, Joel and Desjardins, Guillaume and Rusu, Andrei A and Milan, Kieran and Quan, John and Ramalho, Tiago and Grabska-Barwinska, Agnieszka and others},
  journal={Proceedings of the national academy of sciences},
  volume={114},
  number={13},
  pages={3521--3526},
  year={2017},
  publisher={National Academy of Sciences}
}

@inproceedings{mallya2018packnet,
  title={Packnet: Adding multiple tasks to a single network by iterative pruning},
  author={Mallya, Arun and Lazebnik, Svetlana},
  booktitle={Proceedings of the IEEE conference on Computer Vision and Pattern Recognition},
  pages={7765--7773},
  year={2018}
}

@inproceedings{isele2018selective,
  title={Selective experience replay for lifelong learning},
  author={Isele, David and Cosgun, Akansel},
  booktitle={Proceedings of the AAAI conference on artificial intelligence},
  pages={1},
  year={2018}
}

@article{zhuang2022multi,
  title={Multi-criteria selection of rehearsal samples for continual learning},
  author={Zhuang, Chen and Huang, Shaoli and Cheng, Gong and Ning, Jifeng},
  journal={Pattern Recognition},
  volume={132},
  pages={108907},
  year={2022},
  publisher={Elsevier}
}

@inproceedings{park2019relational,
  title={Relational knowledge distillation},
  author={Park, Wonpyo and Kim, Dongju and Lu, Yan and Cho, Minsu},
  booktitle={Proceedings of the IEEE/CVF conference on computer vision and pattern recognition},
  pages={3967--3976},
  year={2019}
}

@inproceedings{chen2022knowledge,
  title={Knowledge distillation with the reused teacher classifier},
  author={Chen, Defang and Mei, Jian-Ping and Zhang, Hailin and Wang, Can and Feng, Yan and Chen, Chun},
  booktitle={Proceedings of the IEEE/CVF conference on computer vision and pattern recognition},
  pages={11933--11942},
  year={2022}
}

@inproceedings{wang2024improving,
  title={Improving knowledge distillation via regularizing feature direction and norm},
  author={Wang, Yuzhu and Cheng, Lechao and Duan, Manni and Wang, Yongheng and Feng, Zunlei and Kong, Shu},
  booktitle={European Conference on Computer Vision},
  pages={20--37},
  year={2024},
  organization={Springer}
}

@inproceedings{touvron2021training,
  title={Training data-efficient image transformers \& distillation through attention},
  author={Touvron, Hugo and Cord, Matthieu and Douze, Matthijs and Massa, Francisco and Sablayrolles, Alexandre and J{\'e}gou, Herv{\'e}},
  booktitle={International conference on machine learning},
  pages={10347--10357},
  year={2021},
  organization={PMLR}
}

@inproceedings{yang2024vitkd,
  title={Vitkd: Feature-based knowledge distillation for vision transformers},
  author={Yang, Zhendong and Li, Zhe and Zeng, Ailing and Li, Zexian and Yuan, Chun and Li, Yu},
  booktitle={Proceedings of the IEEE/CVF Conference on Computer Vision and Pattern Recognition},
  pages={1379--1388},
  year={2024}
}

@article{luo2025lada,
  title={LADA: Scalable label-specific CLIP adapter for continual learning},
  author={Luo, Mao-Lin and Zhou, Zi-Hao and Wei, Tong and Zhang, Min-Ling},
  journal={arXiv preprint arXiv:2505.23271},
  year={2025}
}

@inproceedings{wang2022learning,
  title={Learning to prompt for continual learning},
  author={Wang, Zifeng and Zhang, Zizhao and Lee, Chen-Yu and Zhang, Han and Sun, Ruoxi and Ren, Xiaoqi and Su, Guolong and Perot, Vincent and Dy, Jennifer and Pfister, Tomas},
  booktitle={Proceedings of the IEEE/CVF conference on computer vision and pattern recognition},
  pages={139--149},
  year={2022}
}

@inproceedings{wang2022dualprompt,
  title={Dualprompt: Complementary prompting for rehearsal-free continual learning},
  author={Wang, Zifeng and Zhang, Zizhao and Ebrahimi, Sayna and Sun, Ruoxi and Zhang, Han and Lee, Chen-Yu and Ren, Xiaoqi and Su, Guolong and Perot, Vincent and Dy, Jennifer and others},
  booktitle={European conference on computer vision},
  pages={631--648},
  year={2022},
  organization={Springer}
}

@inproceedings{smith2023coda,
  title={Coda-prompt: Continual decomposed attention-based prompting for rehearsal-free continual learning},
  author={Smith, James Seale and Karlinsky, Leonid and Gutta, Vyshnavi and Cascante-Bonilla, Paola and Kim, Donghyun and Arbelle, Assaf and Panda, Rameswar and Feris, Rogerio and Kira, Zsolt},
  booktitle={Proceedings of the IEEE/CVF conference on computer vision and pattern recognition},
  pages={11909--11919},
  year={2023}
}

@article{zhang2024continualprompt,
  title={Continual distillation learning: Knowledge distillation in prompt-based continual learning},
  author={Zhang, Qifan and Guo, Yunhui and Xiang, Yu},
  journal={arXiv preprint arXiv:2407.13911},
  year={2024}
}

@inproceedings{li2024promptkd,
  title={Promptkd: Unsupervised prompt distillation for vision-language models},
  author={Li, Zheng and Li, Xiang and Fu, Xinyi and Zhang, Xin and Wang, Weiqiang and Chen, Shuo and Yang, Jian},
  booktitle={Proceedings of the IEEE/CVF Conference on Computer Vision and Pattern Recognition},
  pages={26617--26626},
  year={2024}
}

@inproceedings{zheng2024adapt,
  title={Adapt without forgetting: Distill proximity from dual teachers in vision-language models},
  author={Zheng, Mengyu and Tang, Yehui and Hao, Zhiwei and Han, Kai and Wang, Yunhe and Xu, Chang},
  booktitle={European Conference on Computer Vision},
  pages={109--125},
  year={2024},
  organization={Springer}
}

@inproceedings{zhang2024continual,
  title={Continual Compositional Zero-Shot Learning.},
  author={Zhang, Yang and Feng, Songhe and Yuan, Jiazheng},
  booktitle={IJCAI},
  pages={1724--1732},
  year={2024}
}

@inproceedings{li2026composition,
  title={Composition-incremental learning for compositional generalization},
  author={Li, Zhen and Wu, Yuwei and Jing, Chenchen and Sun, Che and Li, Chuanhao and Jia, Yunde},
  booktitle={Proceedings of the AAAI Conference on Artificial Intelligence},
  volume={40},
  pages={6735--6743},
  year={2026}
}

@inproceedings{li2024context,
  title={Context-based and diversity-driven specificity in compositional zero-shot learning},
  author={Li, Yun and Liu, Zhe and Chen, Hang and Yao, Lina},
  booktitle={Proceedings of the IEEE/CVF Conference on Computer Vision and Pattern Recognition},
  pages={17037--17046},
  year={2024}
}

@inproceedings{isola2015discovering,
  title={Discovering states and transformations in image collections},
  author={Isola, Phillip and Lim, Joseph J and Adelson, Edward H},
  booktitle={Proceedings of the IEEE conference on computer vision and pattern recognition},
  pages={1383--1391},
  year={2015}
}

@inproceedings{yu2014fine,
  title     = {Fine-Grained Visual Comparisons with Local Learning},
  author    = {Yu, Aron and Grauman, Kristen},
  booktitle = {Proceedings of the IEEE Conference on Computer Vision and Pattern Recognition (CVPR)},
  pages     = {192--199},
  year      = {2014}
}

@inproceedings{purushwalkam2019task,
  title={Task-driven modular networks for zero-shot compositional learning},
  author={Purushwalkam, Senthil and Nickel, Maximilian and Gupta, Abhinav and Ranzato, Marc'Aurelio},
  booktitle={Proceedings of the IEEE/CVF international conference on computer vision},
  pages={3593--3602},
  year={2019}
}

@inproceedings{chao2016empirical,
  title={An empirical study and analysis of generalized zero-shot learning for object recognition in the wild},
  author={Chao, Wei-Lun and Changpinyo, Soravit and Gong, Boqing and Sha, Fei},
  booktitle={European conference on computer vision},
  pages={52--68},
  year={2016},
  organization={Springer}
}

@article{kang2025advancing,
  title={Advancing prompt-based methods for replay-independent general continual learning},
  author={Kang, Zhiqi and Wang, Liyuan and Zhang, Xingxing and Alahari, Karteek},
  journal={arXiv preprint arXiv:2503.00677},
  year={2025}
}

@inproceedings{karthik2021revisiting,
  title={Revisiting visual product for compositional zero-shot learning},
  author={Karthik, Shyamgopal and Mancini, Massimiliano and Akata, Zeynep},
  booktitle={NeurIPS 2021 Workshop on Distribution Shifts: Connecting Methods and Applications},
  pages={1},
  year={2021}
}

@InProceedings{Li_2022_CVPR,
    author    = {Li, Xiangyu and Yang, Xu and Wei, Kun and Deng, Cheng and Yang, Muli},
    title     = {Siamese Contrastive Embedding Network for Compositional Zero-Shot Learning},
    booktitle = {Proceedings of the IEEE/CVF Conference on Computer Vision and Pattern Recognition (CVPR)},
    month     = {June},
    year      = {2022},
    pages     = {9326-9335}
}

@InProceedings{Wang_2023_CVPR,
    author    = {Wang, Qingsheng and Liu, Lingqiao and Jing, Chenchen and Chen, Hao and Liang, Guoqiang and Wang, Peng and Shen, Chunhua},
    title     = {Learning Conditional Attributes for Compositional Zero-Shot Learning},
    booktitle = {Proceedings of the IEEE/CVF Conference on Computer Vision and Pattern Recognition (CVPR)},
    month     = {June},
    year      = {2023},
    pages     = {11197-11206}
}

@inproceedings{rebuffi2017icarl,   
    title={icarl: Incremental classifier and representation learning},   
    author={Rebuffi, Sylvestre-Alvise and Kolesnikov, Alexander and Sperl, Georg and Lampert, Christoph H},  
    booktitle={Proceedings of the IEEE conference on Computer Vision and Pattern Recognition},
    pages={2001--2010},
    year={2017} 
}

@article{paszke2019pytorch,
  title={Pytorch: An imperative style, high-performance deep learning library},
  author={Paszke, Adam and Gross, Sam and Massa, Francisco and Lerer, Adam and Bradbury, James and Chanan, Gregory and Killeen, Trevor and Lin, Zeming and Gimelshein, Natalia and Antiga, Luca and others},
  journal={Advances in neural information processing systems},
  volume={32},
  year={2019}
}

@inproceedings{shahapure2020cluster,
  title={Cluster quality analysis using silhouette score},
  author={Shahapure, Ketan Rajshekhar and Nicholas, Charles},
  booktitle={2020 IEEE 7th international conference on data science and advanced analytics (DSAA)},
  pages={747--748},
  year={2020},
  organization={IEEE}
}

\end{document}